%% file: main.tex
\documentclass{article} %
\usepackage{iclr2025_conference,times}

\input{math_commands.tex}

\usepackage{hyperref}
\usepackage{url}
\usepackage{xspace}
\usepackage{enumitem}

\title{WebRL: Training LLM Web Agents via Self-Evolving Online Curriculum Reinforcement Learning}

\author{\centerline{Zehan Qi$^{1*}$, Xiao Liu$^{12*}$, Iat Long Iong$^{1}$, Hanyu Lai$^{1}$, Xueqiao Sun$^{1}$, Wenyi Zhao$^{2}$, Yu Yang$^{2}$}\\
\centerline{\bf Xinyue Yang$^{2}$, Jiadai Sun$^{2}$, Shuntian Yao$^{2}$, Tianjie Zhang$^{2}$, Wei Xu$^{1}$, Jie Tang$^{1}$, Yuxiao Dong$^{1}$}\\\\
\centerline{$^1$Tsinghua University\qquad$^2$Zhipu AI}
}
\iclrfinalcopy

\newcommand{\model}[0]{\textsc{WebRL}\xspace}
\newcommand{\vpara}[1]{\noindent\textbf{#1}\xspace} %

\newcommand{\hide}[1]{} %

\input{preamble}

\begin{document}

\maketitle

\renewcommand{\thefootnote}{\fnsymbol{footnote}}
    \footnotetext[1]{Equal contribution. Emails: \texttt{qzh23@mails.tsinghua.edu.cn}, \texttt{shawliu9@gmail.com}}
    \footnotetext{Work done when ZQ interned at Zhipu AI.}
\renewcommand{\thefootnote}{\arabic{footnote}}

\vspace{-3mm}
\begin{abstract}
Large language models (LLMs) have shown remarkable potential as autonomous agents, particularly in web-based tasks. 
However, existing LLM web agents heavily rely on expensive proprietary LLM APIs, while open LLMs lack the necessary decision-making capabilities. 
This paper introduces \model, a self-evolving online curriculum reinforcement learning framework designed to train high-performance web agents using open LLMs. 
\model addresses three key challenges in building LLM web agents, including the scarcity of training tasks, sparse feedback signals, and policy distribution drift in online learning. 
Specifically, \model incorporates 1) a self-evolving curriculum that generates new tasks from unsuccessful attempts, 2) a robust outcome-supervised reward model (ORM), and 3) adaptive reinforcement learning strategies to ensure consistent improvements. 
We apply \model to transform open Llama-3.1 and GLM-4 models into proficient web agents. %
On WebArena-Lite, \model improves the success rate of Llama-3.1-8B from  4.8\%  to 42.4\%, and from 6.1\% to 43\% for GLM-4-9B. %
These open models significantly surpass the performance of GPT-4-Turbo (17.6\%) and GPT-4o (13.9\%)  and  outperform previous state-of-the-art web agents trained on open LLMs (AutoWebGLM, 18.2\%). 
Our findings demonstrate \model's effectiveness in bridging the gap between open and proprietary LLM-based web agents, paving the way for more accessible and powerful autonomous web interaction systems.
The code, model, and data are made publicly available at \url{https://github.com/THUDM/WebRL}.

\end{abstract}

\hide{
Large language models (LLMs) have shown remarkable potential as autonomous agents, particularly in web-based tasks. 
However, existing LLM web agents face significant limitations: high-performing agents rely on expensive proprietary LLM APIs, while open LLMs lack the necessary decision-making capabilities. 
This paper introduces \model, a self-evolving online curriculum reinforcement learning framework designed to train high-performance web agents using open LLMs. 
Our approach addresses key challenges in building LLM web agents, including the scarcity of training tasks, sparse feedback signals, and policy distribution drift in online learning. 
Specifically, \model incorporates 1) a self-evolving curriculum that generates new tasks from unsuccessful attempts, 2) a robust outcome-supervised reward model (ORM), and 3) adaptive reinforcement learning strategies to ensure consistent improvements. 
We apply \model to transform open Llama-3.1 and GLM-4 models into proficient web agents, achieving remarkable results on the WebArena-Lite benchmark. 
Our Llama-3.1-8B agent improves from an initial 4.8\% success rate to 42.4\%, while the Llama-3.1-70B agent achieves a 49.1\% success rate across five diverse websites. 
These results surpass the performance of GPT-4-Turbo (17.6\%) by over 160\% relatively and significantly outperform previous state-of-the-art web agents trained on open LLMs (AutoWebGLM, 18.2\%). 
Our findings demonstrate \model's effectiveness in bridging the gap between open and proprietary LLM-based web agents, paving the way for more accessible and powerful autonomous web interaction systems.
Code, model, and data will be available at \url{https://github.com/THUDM/WebRL}.
}

\input{section/intro}
\input{section/methodology}

\input{section/experiment}
\input{section/related_work}
\input{section/conclusion}

\textbf{Acknowledgments.}  
We would like to thank Zhipu AI for sponsoring the computation resources and annotation cost used in this work.

\bibliography{iclr2025_conference}
\bibliographystyle{iclr2025_conference}

\appendix
\newpage
\input{section/appendix/algorithm}
\input{section/appendix/baselines}

\input{section/appendix/more_results}
\input{section/appendix/prompts}
\input{section/appendix/qualitive}

\end{document}

%% file: math_commands.tex
\usepackage{amsmath,amsfonts,bm}

\def\eqref#1{equation~\ref{#1}}

\def\1{\bm{1}}

\DeclareMathAlphabet{\mathsfit}{\encodingdefault}{\sfdefault}{m}{sl}
\SetMathAlphabet{\mathsfit}{bold}{\encodingdefault}{\sfdefault}{bx}{n}

\newcommand{\E}{\mathbb{E}}

\DeclareMathOperator*{\argmax}{arg\,max}
\DeclareMathOperator*{\argmin}{arg\,min}

%% file: preamble.tex
\usepackage{hyperref}
\usepackage[all]{hypcap}

\usepackage{tcolorbox}
\usepackage{amsmath}
\usepackage{amssymb}
\usepackage{mathtools}
\usepackage{booktabs}
\usepackage{threeparttable}
\usepackage{tabularx}
\usepackage{xspace}
\usepackage{soul}
\usepackage{xcolor}
\usepackage{subfig}

\usepackage{wrapfig}  %
\usepackage{lipsum}   %

\usepackage{graphicx}
\usepackage{caption}
\usepackage{subcaption}
\usepackage{subfig}      %

\usepackage{bm} %

\usepackage{algorithm}
\usepackage{algorithmic}

\usepackage{multirow}

\newcommand{\ie}{{\em i.e.}}

\newenvironment{packeditemize}{
\begin{list}{$\bullet$}{
\setlength{\labelwidth}{8pt}
\setlength{\itemsep}{0pt}
\setlength{\leftmargin}{\labelwidth}
\addtolength{\leftmargin}{\labelsep}
\setlength{\parindent}{0pt}
\setlength{\listparindent}{\parindent}
\setlength{\parsep}{3pt}
\setlength{\topsep}{3pt}}}{\end{list}}

\usepackage{pifont}

\tcbset{
    mybox/.style={
        fontupper=\linespread{.9}\selectfont,
        top=0pt, %
        bottom=0pt %
    }
}
\newcommand{\orm}{$\mathcal{M}_{\text{ORM}}$}

\definecolor{mygreen}{RGB}{25, 190, 39}
\definecolor{myblue}{RGB}{46, 128, 208}
\definecolor{myorange}{RGB}{230, 96, 6}
\definecolor{myred}{RGB}{200, 29, 49}

%% file: section/intro.tex
\begin{figure}[htbp]
    \vspace{-6mm}
    \centering\subfloat[Performance comparison between proprietary LLMs and open-sourced LLMs on WebArena-Lite.]{
        \includegraphics[width=0.48\linewidth]{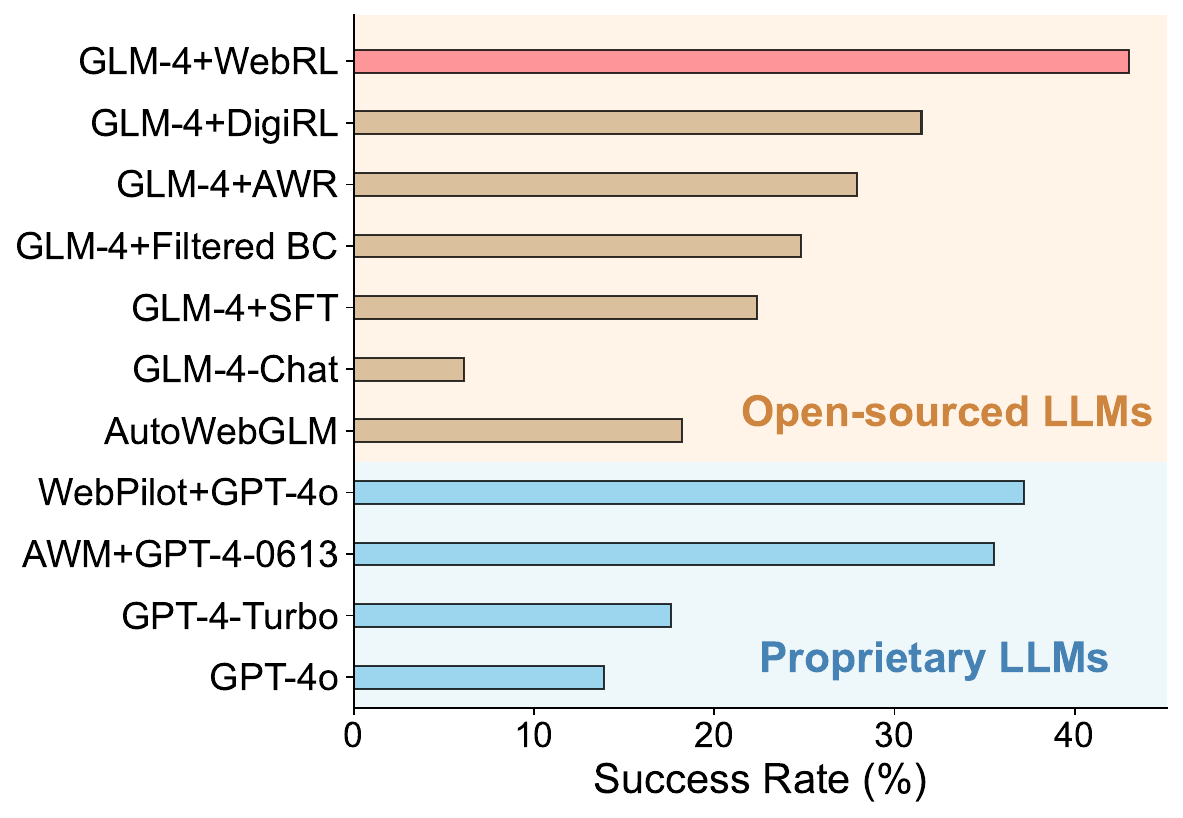}
    }
    \hspace{0.01\linewidth}  %
    \subfloat[Performance changes of GLM-4-9B trained with \model and baseline methods.]{
        \includegraphics[width=0.4\linewidth]{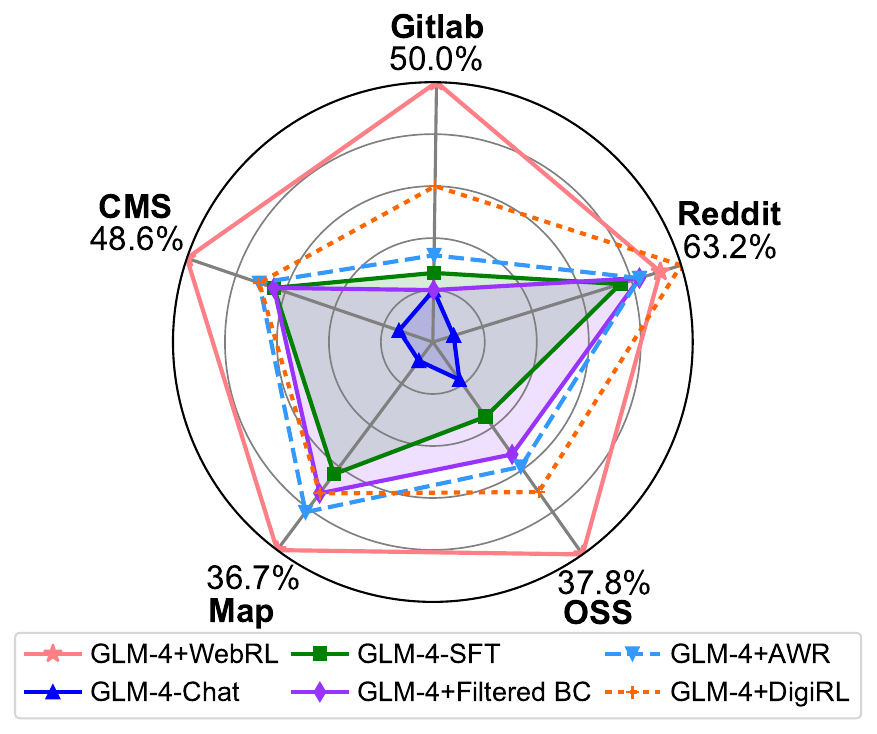}
    }
    \caption{(a) Compared with all proprietary and open-sourced LLMs, GLM-4-9B with \model achieves the best results. (b) The performance of GLM-4-9B on WebArena-Lite~\citep{zhou2023webarena,liu2024visualagentbench}, trained using \model, shows significant improvement over other baselines across all five evaluated websites.}
    \label{fig:two_subfigs}
\end{figure}

\section{Introduction}
Large language models (LLMs) have exhibited not only superior comprehension of human language, commonsense reasoning, and knowledge acquisition, but also significant potential in complex planning and logical reasoning, indicating their promising trajectory towards serving as autonomous LLM agents~\citep{wang2023voyager, liu2023agentbench}. 
A diverse array of applications for LLM agents has proliferated, encompassing domains such as code generation~\citep{jimenez2024swebench}, database manipulation~\citep{zhou2023llm,gu2024middleware}, and graphical user interface (GUI) interaction~\citep{rawles2024androidinthewild,yang2023appagent,xie2024osworld}. 
Among these, web agents powered by LLMs~\citep{deng2024mind2web,zheng2024gpt,lai2024autowebglm,pan2024autonomous} have garnered particular attention due to their extensive application prospects and unique potential for fostering authentic autonomous intelligence within the digital ecosystem.

Notwithstanding these advancements, existing LLM web agents, regardless of their performance metrics or architectural paradigms, remain under-developed. 
High-performing LLM web agents predominantly rely on meticulously crafted prompts in conjunction with proprietary LLM APIs (e.g., OpenAI GPT-4) for web page comprehension and manipulation, which is both expensive and time-intensive. 
Conversely, open-source LLMs exhibit notable deficiencies in their capability to function as proficient web agents, primarily due to the scarcity of decision-centric data in both pre-training and post-training periods. 
Despite recent endeavors~\citep{lai2024autowebglm,pan2024autonomous} to train web agents on open LLMs via imitation learning, these approaches insufficiently leverage the inherently online nature of web interactions and fail to yield consistent, continual improvements.

\vpara{Challenges.}
In this work, we propose to train high-performance web agents based on open LLMs within online environments, specifically utilizing WebArena~\citep{zhou2023webarena}. 
Our investigation has identified several critical challenges inherent to this task:
1) \textit{Insufficiency of training tasks}: In contrast to offline datasets~\citep{deng2024mind2web,rawles2024androidinthewild} that facilitate agent training and evaluation on human-annotated oracle trajectories, online benchmarks such as WebArena typically provide only a limited test set for evaluation purposes. 
This dearth of predefined training tasks significantly impedes the effective training of agents within these environments.
2) \textit{Sparsity and cost of feedback signals}: The assessment of success for arbitrary web browsing tasks is difficult in the absence of task-specific evaluation functions. 
Moreover, unlike tasks in certain GUI datasets (e.g., AITW~\citep{rawles2024androidinthewild} and WebShop~\citep{yao2022webshop}), those in WebArena are typically of long horizons, with oracle solutions averaging about 10 steps. This characteristic introduces substantial sparsity in the available signals during online exploration.
3) \textit{Policy distribution drift in online learning}: The absence of a predefined training set necessitates online exploration, inevitably leading to distribution drift in the agent's policy. 
This phenomenon is likely to induce catastrophic forgetting and performance degradation over time.

\vpara{The \model Framework.}
In response to these challenges, we introduce \model, a self-evolving online curriculum reinforcement learning framework designed for training LLM web agents. 
To the best of our knowledge, this represents the first systematic framework enabling effective reinforcement learning for LLM web agents from initialization in online web environments. 
Through the application of \model, we have successfully transformed a Llama-3.1-8B model into a proficient LLM web agent, elevating its success rate (SR) on WebArena-Lite~\citep{zhou2023webarena,liu2024visualagentbench} from an initial 4.8\% to 42.4\% across a diverse set of five websites. 
Furthermore, when applied to Llama-3.1-70B, we achieve a remarkable 49.1\% SR, significantly surpassing the performance of the most advanced proprietary LLM API (GPT-4-Turbo, 17.6\% SR) and the previous state-of-the-art web agents trained on open-source LLMs (AutoWebGLM~\citep{lai2024autowebglm}, 18.2\% SR).

\begin{figure}[t]
    \centering
    \includegraphics[width=\linewidth]{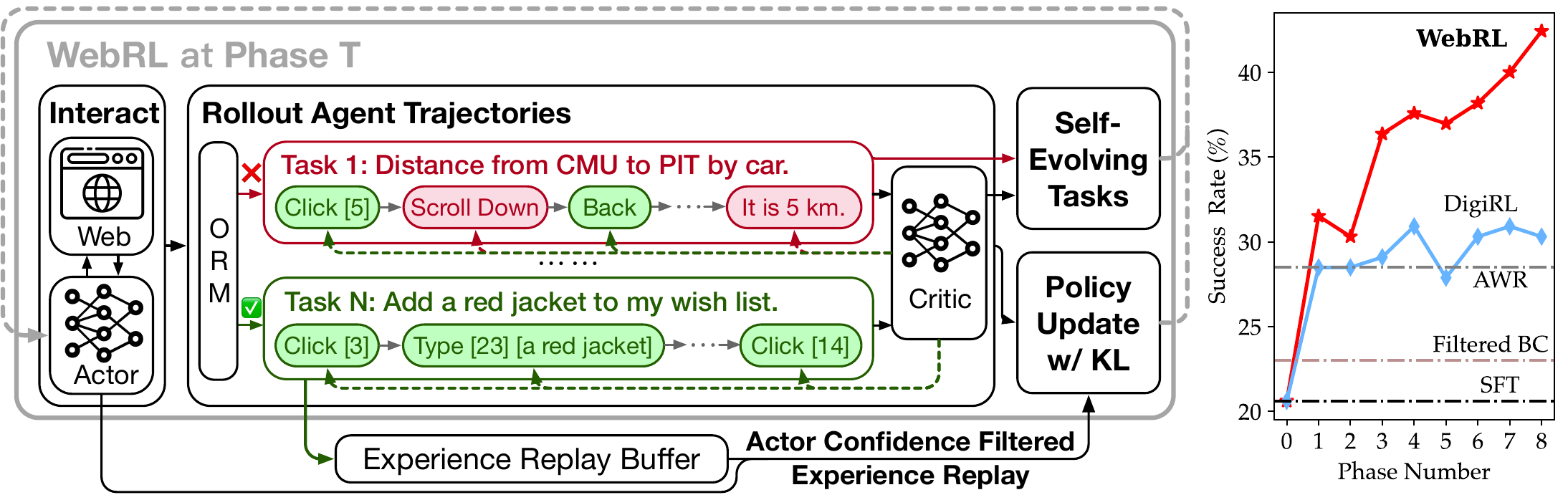}%
    \caption{Overview of \model. \model is a self-evolving online curriculum reinforcement learning framework for LLM-based web agents, yielding consistent continual improvements throughout the iterative self-evolution.}
    \label{fig:framework}
    \vspace{-5mm}
\end{figure}

The substantial performance gains from \model can be attributed to several key architectural designs.
To address the scarcity of web agent training tasks, we have devised a self-evolving online curriculum that harnesses the trial-and-error process inherent in exploration. 
This curriculum is underpinned by a robust outcome-supervised reward model (ORM) that we have newly developed. 
In each training phase, novel tasks are autonomously generated from unsuccessful attempts in the preceding phase, facilitating a progressive learning trajectory.
To mitigate the policy distribution shift induced by curriculum-based reinforcement learning, we incorporate a KL-divergence term between the reference and actor policies into our learning algorithm, thereby constraining policy updates and promoting stability.
We implement an experience replay buffer augmented with a novel actor confidence filtering strategy to ensure the fidelity of replayed experiences and prevent over-fitting to previously acquired knowledge. 
The experimental results confirm the effectiveness of \model. In particular, the agent demonstrates improved performance when selecting past experiences of moderate difficulty—neither too simple nor too challenging relative to the agent's current capabilities. Additionally, the use of a larger KL divergence constraint in the policy update process results in better performance when incorporating past experience.

In summary, our work makes the following significant contributions to the field:
\begin{itemize}[leftmargin=1.5em,itemsep=0pt,parsep=0.2em,topsep=0.0em,partopsep=0.0em]
\item We introduce \model, a novel self-evolving online curriculum RL framework for training LLM-based web agents. For the first time, it implements the infrastructure for RL in the WebArena environment, together with a strong ORM, to drive open LLMs to become capable web agents.
\item \model advances the RL for LLM agent training by addressing key challenges including the scarcity of training tasks, sparsity of feedback signals, and distribution drift in online learning. The self-evolving curriculum and adaptive learning strategies allow the consistent continual improvement of LLM web agents during iteration.
\item We demonstrate \model's substantial performance improvements over existing methodologies such as AWR and DigiRL, achieving state-of-the-art results on the WebArena-Lite benchmark. It surpasses the best proprietary LLM API and previously trained web agent on open LLMs by over 160\% relatively.
\end{itemize}

%% file: section/methodology.tex
\section{\model: Self-Evolving Online Curriculum RL}

We present a self-evolving online curriculum learning framework designed for training web agents, targeting the WebArena~\citep{zhou2023webarena} environment. 
In this system, as illustrated in Figure~\ref{fig:framework}, the agent continuously interacts with its environment to collect real-time trajectory data. This interaction is guided by the self-evolving curriculum learning strategy that dynamically generates tasks, effectively mitigating the insufficiency of training tasks. Furthermore, the tasks generated by the self-evolving curriculum learning strategy are tailored to the agent's current proficiency, thereby increasing the likelihood of receiving positive feedback and alleviating the challenge of sparse feedback signals. Additionally, we train an outcome-supervised reward model (ORM) to evaluate task success. We introduce a KL-constrained policy update algorithm that prevents severe policy shifts during curriculum learning. A replay buffer is also utilized to retain prior knowledge and mitigate the risks of catastrophic forgetting. These techniques enable the agent to improve incrementally, progressively handling more complex tasks. The overall training process can be found in Algorithm~\ref{algorithm:1}.

\textbf{Problem Formulation.} 
We model the process of completing the web task as a finite-horizon Markov Decision Process (MDP), denoted by \((S, A, R,\mathcal{T})\). 
Given a user instruction \(I\), the agent is required to complete the corresponding task. 
The state \(s\) is defined as the HTML content of the current web page along with the history of previous actions. 
The agent receives a reward of 1 upon successful task completion, and 0 otherwise. 
In the finite-horizon setting, the trajectory ends either when the task is accomplished or when the maximum number of interactions \(T\) is exceeded. 
To explain our method clearly, we introduce the following notation. 
The policy \(\pi(\cdot | s_t, I)\) represents the distribution over actions given the state \(s_t\) and the instruction \(I\). 
The value function \(V(s_h, I) = \mathbb{E}_{\pi} \left[ \sum_{t=h}^T r(s_t, a_t, I) \right]\) represents the expected cumulative reward from the state \(s_h\) under policy \(\pi\). 
The action-value function $Q(s_t, a_t, I)$ is the expected cumulative reward for taking action $a_t$ on state $s_t$ and following policy $\pi$ thereafter: $Q(s_t, a_t, I) = r(s_t, a_t) + V(s_{t+1}, I)$.

\vpara{ORM Training.}
In the curriculum learning process, we need to determine whether the corresponding instruction is completed based on the trajectory generated by the agent. 
Due to the lack of feedback from the environment, we train an LLM as the outcome-supervised reward model \orm{} to achieve this task success evaluation. 
\orm{} is utilized to assess whether the agent's rollout trajectory accomplishes a given task, providing a binary reward signal (0 for failure and 1 for success).

Similar to the approach in \citep{zhang2024generative}, we configure \orm{} to output ``YES'' or ``NO'' to indicate whether a trajectory successfully completes a task, leveraging the learned knowledge from the language head of \orm{}. 
Given the limited context window of LLMs and the typically long length of HTML documents, we adopt a strategy akin to ~\citep{pan2024autonomous}, keeping the HTML of only the final state to the input. 
In addition, the historical actions of agents, which provide information about previous steps of trajectories are also included.
Thus, the input to the model consists of several components: the instruction \(I\), historical actions, and HTML of the final state. 
We wrap these components into the prompt asking the model to determine whether the trajectory successfully completes the task described by instruction \(I\). 
To obtain the outcome, we compare the probabilities of generating ``YES'' and ``NO'' from \orm{}. 
If the probability of generating ``YES'' is higher than that of generating ``NO'', the task is considered successful, and the reward is set to 1. 
Otherwise, the reward is set to 0.

\subsection{Self-evolving New Instruction for Curriculum Learning}

A typical challenge in training LLM web agents within WebArena is the scarcity of training tasks, resonating with the situation of developing real-world web agents.
Although the recent work~\citep{liu2024visualagentbench} has curated a trajectory fine-tuning set for WebArena, it only contains around 1k instructions with oracle trajectories, far from enough for training strong LLM web agents.
To address this limitation and drive continuous improvement, we employ a self-evolving curriculum learning strategy. This method generates new training instructions at each phase. As the phase progresses, the generated instructions become increasingly complex, allowing the agent's capabilities to improve gradually. We implement a two-step process of generation and filtering, to produce tasks that are incrementally more challenging, while still being suitable for the agent's current capability. During the generation step, we use the in-breadth evolving approach~\citep{xu2023wizardlm} to create new instructions. We select instructions the model failed to complete in previous interaction phases as seeds for generating new instructions. Detailed prompts are provided in the Appendix~\autoref{sec:prompts}.
To ensure that the generated instructions are both feasible in the target environment and aligned with the desired difficulty level, we first filter them using the trained critic. Specifically, we use the critic to evaluate each new instruction by considering its initial state. We select instructions with critic scores between 0.05 and 0.75, ensuring that only tasks meeting our difficulty criteria are retained. 
\textcolor{black}{We manually review generated tasks and identify tasks that cannot be completed in WebArena. Based on these findings, we develop a prompt (Figure~\ref{fig:prompt_for_filter}) and use GPT-4o to exclude infeasible tasks in WebArena automatically.}
The resulting set of instructions is used for interaction and training in this phase.

\subsection{Reinforcement Learning for LLMs in Online Web Environments}

In each phase of curriculum learning, the model progressively encounters and learns a new set of tasks. 
Considering this setting, a major challenge here is to avoid excessive policy distribution drift during each learning phase, which could lead to the catastrophic forgetting of previously acquired knowledge. 
Traditional approaches typically mitigate the issue by mixing data from different phases. 
However, in web agent tasks, intermediate steps do not receive direct process rewards, with only weak signals from the outcome of the final state. 
Consequently, even if an intermediate step is executed correctly, an error in later steps can easily lead to the final failure, resulting in misjudgment of the intermediate step and making it difficult to be reused.
As a result, in this work, we primarily seek algorithmic improvements to address policy distribution drift more directly.

A potential solution comes from ideas in reinforcement learning with human feedback (RLHF)~\citep{ouyang2022training}, where the Kullback-Leibler (KL) divergence between two policies is constrained to mitigate policy distribution drift. By adapting this to our curriculum learning setup, we aim to ensure that the policy in the current phase does not deviate too much from the policy in the previous phase, while still optimizing performance on new tasks. Let the policy from the previous phase be denoted as \( \pi_{\text{ref}} \), and the current policy being optimized as \( \pi_\theta \). The instruction distribution for the current phase is represented as \( \rho(I) \). The objective for optimizing \( \pi_\theta \) in the current phase can then be written as follows:
\begin{equation} \label{eq: target}
    \max_{\pi_\theta} \E_{I \sim \rho(I), a_t \sim \pi_\theta(\cdot|s_t)} \left[ \sum_{t=0}^{T} \left( r(s_t, a_t, I) + \beta \log \pi_{\text{ref}}(a_t|s_t, I) \right) + \beta \mathcal{H}(\pi_\theta) \right]
\end{equation}
where  \( \beta \) is a coefficient controlling the strength of the KL divergence constraint and \( \mathcal{H}(\pi_\theta) \) represents the entropy of the current policy.

Following the work of \citep{rafailov2024r}, we can interpret the objective of eq.~\ref{eq: target} as a maximum entropy reinforcement learning problem. The optimal policy $\pi^*$ for this problem can be expressed as:
\begin{equation} \label{eq: optimal_pi}
    \pi^*(a_t | s_t, I) = e^{\left(Q^*(s_t, a_t, I) - V^*(s_t, I)\right)/\beta}
\end{equation}
where \( V^*(s_t, I) \) is the optimal value function, representing the expected cumulative reward under the optimal policy \( \pi^* \). \( Q^*(s_t, a_t, I) \) is the optimal action-value function. The relationship between \( Q^* \) and \( V^* \) is given by:
\begin{equation} \label{eq: Q_V_relation}
    Q^*(s_t, a_t, I) = 
\begin{cases} 
r(s_t, a_t, I) + \beta \log \pi_{\text{ref}}(a_t | s_t, I) + V^*(s_{t+1}, I), & \text{if } s_{t+1} \text{ is not terminal} \\
r(s_t, a_t, I) + \beta \log \pi_{\text{ref}}(a_t | s_t, I), & \text{if } s_{t+1} \text{ is terminal}
\end{cases}
\end{equation}
Based on eq.~\ref{eq: optimal_pi} and eq.~\ref{eq: Q_V_relation}, we can derive:
\begin{equation}\label{eq:loss}
    \beta \log \frac{\pi^*(a_t|s_t, I)}{\pi_{\text{ref}}(a_t|s_t, I)} = r(s_t, a_t, I) + V^*(s_{t+1}, I) - V^*(s_t, I) = A^*(s_t, a_t, I)
\end{equation}
Here, \( A^*(s_t, a_t, I) \) indicates the advantage of taking action \( a_t \) in state \( s_t \) compared to the average reward expected in that state.
Based on the condition, we can formulate the loss function of policy $\pi_\theta$ as:
\begin{equation} \label{eq: loss_func}
    \mathcal{L}(\pi_\theta) = \E_{\nu} \left[ \left( \beta \log \frac{\pi_\theta(a|s, I)}{\pi_{\text{ref}}(a|s, I)} - A^*(s, a, I) \right)^2 \right]
\end{equation}
where $\nu(s)$ represents the distribution of experience in this phase. Note that our algorithm operates in the off-policy manner. A comprehensive derivation, thorough analysis, and a detailed comparison with other RL algorithms can be found in Appendix~\ref{sec:derivation}.

\textbf{What Does The Update Do? } To gain a mechanistic understanding of the loss function, we analyze the gradient of the loss function, \( \mathcal{L}(\pi_\theta) \). The gradient with respect to the parameters \( \theta \) can be expressed as:
\begin{equation}
    \nabla_\theta \mathcal{L}(\pi_\theta) = -2\beta \E_{\nu} \Big[  
    \underbrace{\big(A^*(s, a, I)\vphantom{\frac{\pi_\theta(a|s, I)}{\pi_{\text{ref}}(a|s, I}}}_{\smash{\text{update direction}}} - \underbrace{\beta \log \frac{\pi_\theta(a|s, I)}{\pi_{\text{ref}}(a|s, I)}}_{\smash{\text{KL divergence constraint}}}\big) 
     \underbrace{\nabla_\theta \log\pi_\theta(a|s,I)\vphantom{\frac{\pi_\theta(a|s, I)}{\pi_{\text{ref}}(a|s, I}}}_{\smash{\text{sft loss}}} 
     \Big]
\end{equation}
The gradient demonstrates the following attributions:
\begin{packeditemize}
    \item When the advantage \( A^*(s, a, I) > 0 \), action \( a \) is valuable, so its probability should increase. If \( \pi_\theta \) is lower than \( \pi_\text{ref} \), this increase will be amplified, especially as the gap between them grows. If \( \pi_\theta \) is already higher than \( \pi_\text{ref} \), the increase will be moderated to avoid excessive deviation.
  
    \item When \( A^*(s, a, I) < 0 \), the action is suboptimal, so its probability should decrease. If \( \pi_\theta \) is lower than \( \pi_\text{ref} \), the KL divergence constraint will limit how much it can be reduced to avoid a large divergence. If \( \pi_\theta \) is higher than \( \pi_\text{ref} \), a larger decrease will be allowed.

    \item The parameter \( \beta \) controls the strength of the KL divergence constraint. Adjusting \( \beta \) can help fine-tune this constraint. For instance, increasing \( \beta \) can prevent unnecessary boosts in action probabilities when \( \pi_\text{ref} \) already assigns a high probability to an action.
\end{packeditemize}

\textbf{Training a Reliable Advantage Estimator.}
A reliable advantage estimator is essential for effective policy updates. We train a value network \(V(s_t, I)\) and use Generalized Advantage Estimation (GAE)~\citep{schulman2015high} to compute the advantage. In our setting, we only receive a binary reward (0 or 1) at the final step, with no intermediate rewards (\ie{}, intermediate rewards are effectively zero). Following recent approaches~\citep{farebrother2024stop}, we train the value network using a cross-entropy objective. The loss function for the value network \(V\) is defined as:
\begin{equation} \label{eq:loss_v}
    \mathcal{L}(V) = -\E_{\nu} \Big[ r(s_T, a_T, I) \log V(s, a, I) 
    + (1 - r(s_T, a_T, I)) \log (1 - V(s, a, I)) \Big]
\end{equation}
In line with~\citep{bai2024digirl}, we focus solely on the next-step and final-step advantage estimators, since there is no intermediate reward.
\begin{equation} \label{eq:gae}
    A(s_t, a_t, I) = \lambda \big(r(s_t, a_t, I) + V(s_{t+1}, I) - V(s_t, I)\big) + (1 - \lambda) \big(r(s_T, a_T, I) - V(s_t, I)\big)
\end{equation}
where $\lambda$ is a balancing factor that controls the trade-off between bias and variance in advantage estimation. We set $\lambda$ as 0.5 in our work. 
We approximate the true advantage function \( A^* \) using the estimated advantage function \( A \), which is derived from the value network \( V \). The feasibility of this approximation is demonstrated in Appendix~\ref{sec:critic}, where we show that using \( A \) to update the policy can lead to policy improvement.

\textbf{Experience Replay Buffer with Actor Confidence Filtering.} 
In addition to controlling the policy distribution drift at the algorithmic level through KL divergence, we also implement an adaptive replay buffer to alleviate knowledge forgetting at the data level. 
Specifically, we only store those successful trajectories (which can be sparse) from each phase in the replay buffer. 
During phase $i$, we use the actor from the last phase to compute the perplexity of all actions in the buffer. Actions with a perplexity within the range of 1/0.95 to 1/0.5, along with their corresponding states, are added to the training data for the current phase. This filtering process excludes both over-familiar data and data that remains too challenging for the actor. Additionally, by storing only successful trajectories, we avoid the challenge of accurately estimating intermediate states for incorrect trajectories from previous phases.

%% file: section/experiment.tex
\section{Experiments}

\subsection{Environments and Baselines}

\textbf{Environments.} The effectiveness of \model and baseline methods is evaluated using the WebArena environment~\citep{zhou2023webarena}. WebArena is particularly well-suited to our needs, as it provides a highly interactive platform that supports online learning. Additionally, WebArena encompasses a variety of websites, including OpenStreetMap (Map), Reddit, GitLab, online store content management system (CMS), and OneStopShop (OSS), making it an ideal benchmark for comprehensively assessing model performance on web tasks.
In the original WebArena environment, a total of 812 instructions are provided. Considering the cost of testing, we use 165 test cases from WebArena-Lite~\citep{liu2024visualagentbench} for evaluation.

\textbf{Baselines. }
We compare \model with proprietary LLMs utilizing prompting techniques, as well as open-sourced LLMs trained with alternative methods. For proprietary models, we select GPT-4-Turbo-2024-0409 (GPT-4-Turbo)~\citep{achiam2023gpt} and \href{https://openai.com/index/hello-gpt-4o/}{GPT-4o}. In addition to AWM~\citep{wang2024agent} and WebPilot~\citep{zhang2024webpilot}, we also use the results of models under the simple prompt as baselines. Details of the simple prompt can be seen in Appendix~\autoref{sec:prompts}.
For the open-source models, in addition to using these models with the simple prompt as baselines, we also train Llama3.1~\citep{dubey2024llama} and GLM-4-9B~\citep{glm2024chatglm} using various approaches as baselines. Specifically, we employ imitation learning, also referred to as supervised fine-tuning (SFT), to train these models. The training data is derived from publicly available human-labeled demonstrations, sourced from the WebArena-Lite. 
In addition, we also explore several reinforcement learning methods for comparison, including Filtered Behavior Cloning (Filtered BC)~\citep{pan2024autonomous}, advantage-weighted regression (AWR)~\citep{peng2019advantage} and DigiRL~\citep{bai2024digirl}. For \model and the reinforcement learning-based baselines, we utilize the SFT-trained model as the initial model for the actor. The critic is similarly based on the SFT-trained model, with the addition of a randomly initialized value head. The training details of \model and baselines can be found in Appendix~\autoref{sec:training}.

\input{table/main_result}

\textbf{ORM. }
WebArena-Lite~\citep{liu2024visualagentbench} provides training samples along with a corresponding reward function. We further enhance this set of data by introducing task rewrites, as well as modifying certain data variables, such as place names and product names. We also make adjustments to the associated reward function. \orm{} is trained using rollouts of \model and part of baseline methods on this set of tasks, with evaluation results determined by the reward function. More details can be found in Appendix~\autoref{sec:training}.

\subsection{Main Results}
Our main results, presented in Table~\ref{tab:main_result}, show that Llama3.1-8B trained using \model achieves an average accuracy of 42.4\%, surpassing all baselines, including prompting and training alternatives. Notably, \model excels in specific tasks such as Gitlab (46.7\%) and CMS (54.3\%), demonstrating its ability to address complex web tasks effectively.
Reinforcement learning-based approaches outperform those based on imitation learning, including SFT and Filtered BC, which tend to over-repeat certain actions. For instance, in the table analysis task of CMS, SFT-trained models often over-optimize the ``Scroll Down'' action, which occurs with high frequency. This over-optimize can cause the model to become
trapped in local loops, thereby hindering its ability to achieve the overall task objective effectively. In contrast, reinforcement learning mitigates this by using a critic to estimate the value of each step, optimizing for long-term cumulative rewards, hence enabling more effective handling of complex, multi-step tasks. Furthermore, \model consistently outperforms DigiRL. A significant limitation of DigiRL is that it conducts policy updates on a predefined, fixed set of tasks, which may not align with the model's current skill level. Some of these tasks are particularly challenging for the model to learn due to the sparse reward situations.
This misalignment can cause the model
to converge to suboptimal solutions and restrict its capacity for exploration and skill advancement. \model addresses this limitation by employing self-evolving curriculum learning, adjusting the task complexity based on the model's current abilities. This strategy promotes wider exploration and supports continuous improvement. A similar phenomenon is also observed in the case of the GLM-4-9B, providing evidence that the benefits of \model extend across different model architectures, validating its robustness and adaptability.

\subsection{Scaling Effect of \model}

We further validate the effectiveness of \model on larger-scale models by training Llama3.1-70B using \model. The specific results are presented in Table~\ref{tab:main_result}. After training with \model, Llama3.1-70B achieves an overall accuracy of 49.1\%, reflecting a 26.1\% improvement over the accuracy achieved with SFT. This indicates that \model is scalable and can be effectively applied to larger-scale models.
Furthermore, when comparing the performance improvement from Llama3.1-8B to Llama3.1-70B achieved through SFT, \model demonstrates even greater performance gains as the model scale increases.

\subsection{Distribution Analysis of Error Types}

\begin{figure}[t]
    \centering
    \includegraphics[width=0.95\linewidth]{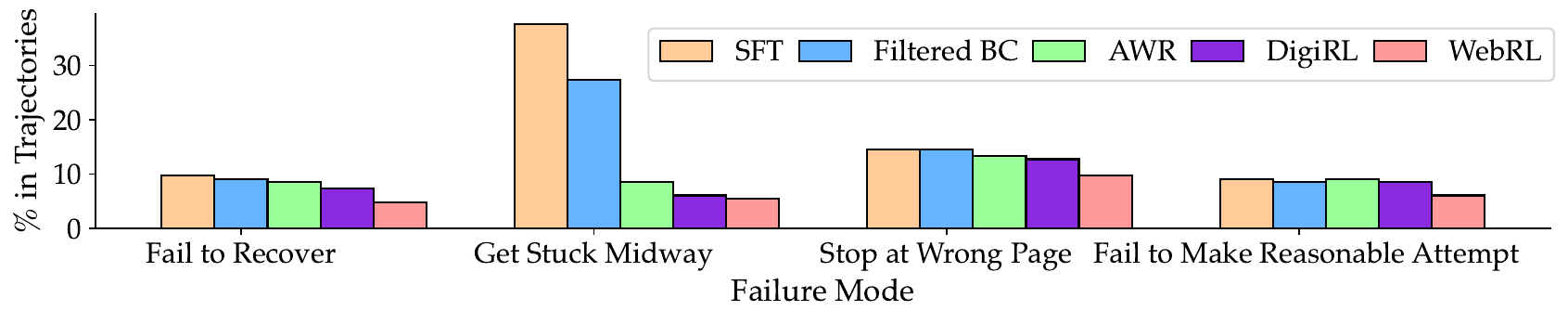}
    \caption{Distribution analysis of error types for \model and baseline methods.}
    \label{fig:error_analysis}
\end{figure}

We compare the performance of Llama 3.1-8B trained with \model against baseline methods across different error types: ``Fail to Recover", ``Get Stuck Midway'', ``Stop at Wrong Page'', and ``Fail to Make Reasonable Attempt'', as shown in Figure~\ref{fig:error_analysis}.
\model demonstrates significant advantages in reducing the ``Get Stuck Midway'' error, especially compared to SFT and Filtered BC. The ``Get Stuck Midway'' error typically arises when the model gets trapped in a loop, repeatedly executing the same action without making progress. Reinforcement learning helps mitigate this issue by optimizing each action while considering its overall impact on the task, enabling the model to make more effective decisions.
Additionally, the model trained with \model demonstrates enhanced robustness in handling the ``Fail to Recover'' error. Through curriculum learning, the model gradually learns how to adapt its actions when encountering failures. For example, when the search query ``Pharmacy near CMU within a 20-minute walking distance" does not yield the desired results, the model learns to modify the query to ``Pharmacy near CMU" and attempts the search again, rather than repeating ineffective actions.
In addition, \model exhibits the lowest error rate on both ``Stop at Wrong Page'' and ``Fail to Make Reasonable Attempt'' errors, indicating the model trained with \model has a more profound comprehension of the relationship between tasks and web pages. It can better identify the correct page needed to complete a specific task, reducing the chances of mistakenly stopping on the wrong page or navigating to an incorrect page. 

\subsection{Performance on Tasks with Varying Step Requirements}

\begin{figure}[t]
  \centering
  \begin{minipage}[t]{0.43\textwidth}
    \centering
    \includegraphics[width=1.0\linewidth]{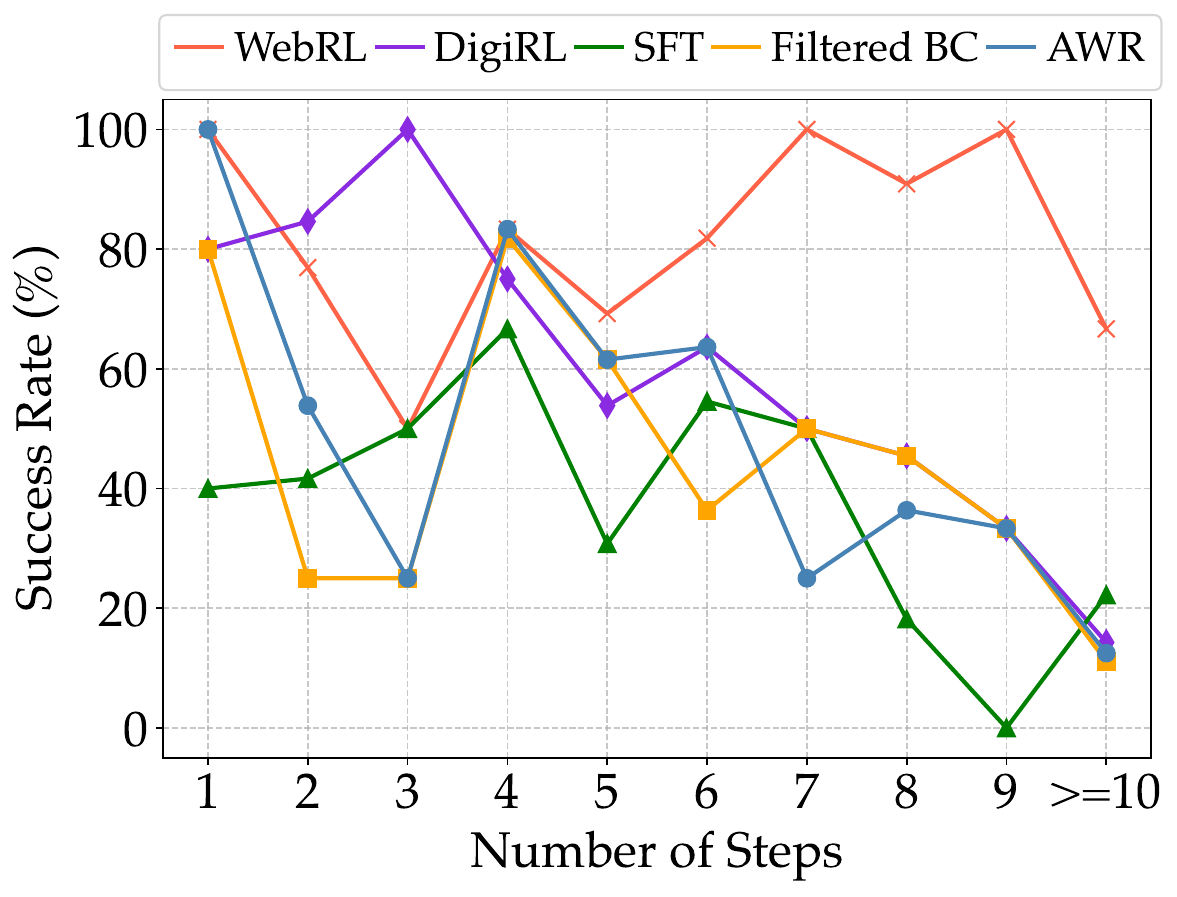}
    \vspace{-5mm}
    \caption{Accuracy of \model and baselines for tasks requiring different steps.}
    \label{fig:step}
  \end{minipage}
  \hfill
  \begin{minipage}[t]{0.55\textwidth}
    \centering
    \includegraphics[width=\linewidth]{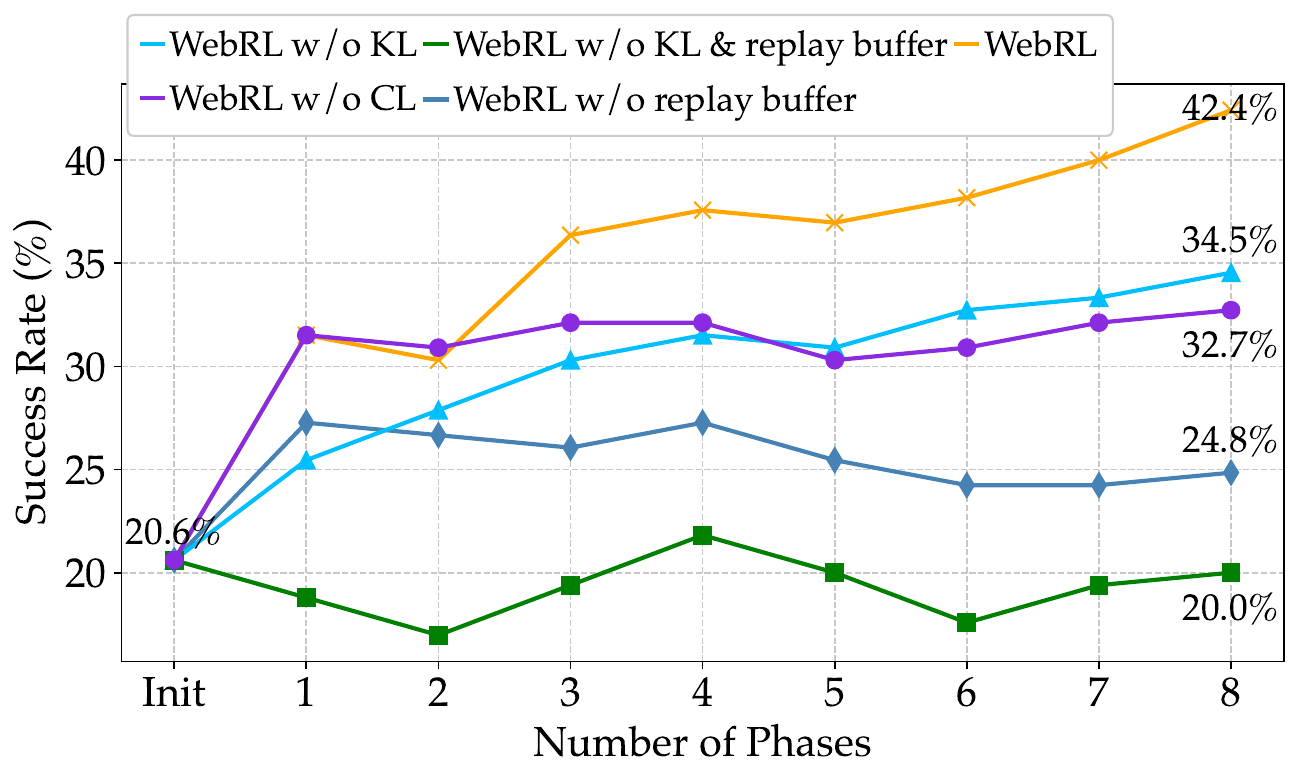}
    \vspace{-5mm}
    \caption{Ablation study of \model on replay buffer, KL-constrained policy update and curriculum strategy.}
    \label{fig:ablation}
  \end{minipage}
  \vspace{-5mm}
\end{figure}

We evaluate the performance of Llama3.1-8B, trained using \model and baseline methods, on tasks with varying step requirements. To determine the required step count for each task, we exclude tasks that no model completes and use the trajectory with the fewest steps as the required step count for each remaining task. The results are shown in Figure~\ref{fig:step}.
It can be seen that the performance of models trained with SFT and Filtered BC shows a noticeable decline as the task length increases. This is likely because these models optimize individual steps without considering the cumulative impact, making them less effective on long-horizon tasks. DigiRL-trained model improves performance on medium-length tasks but struggles with longer tasks (more than 10 steps). This limitation may stem from DigiRL's online learning on a fixed set of tasks. Even when the model executes intermediate steps correctly, it doesn't receive positive rewards if errors occur in later steps, making it harder for the model to learn how to complete tasks that require many steps effectively.
In contrast, \model overcomes this issue with curriculum learning, progressively increasing task difficulty. This approach enhances the model's ability to handle long sequences, leading to significant performance improvements on tasks requiring long-term planning compared to other methods.

\subsection{Performance on Tasks with Varying Complexity}

\begin{wrapfigure}{r}{0.35\textwidth}
    \centering
    \vspace{-4mm}
    \includegraphics[width=1\linewidth]{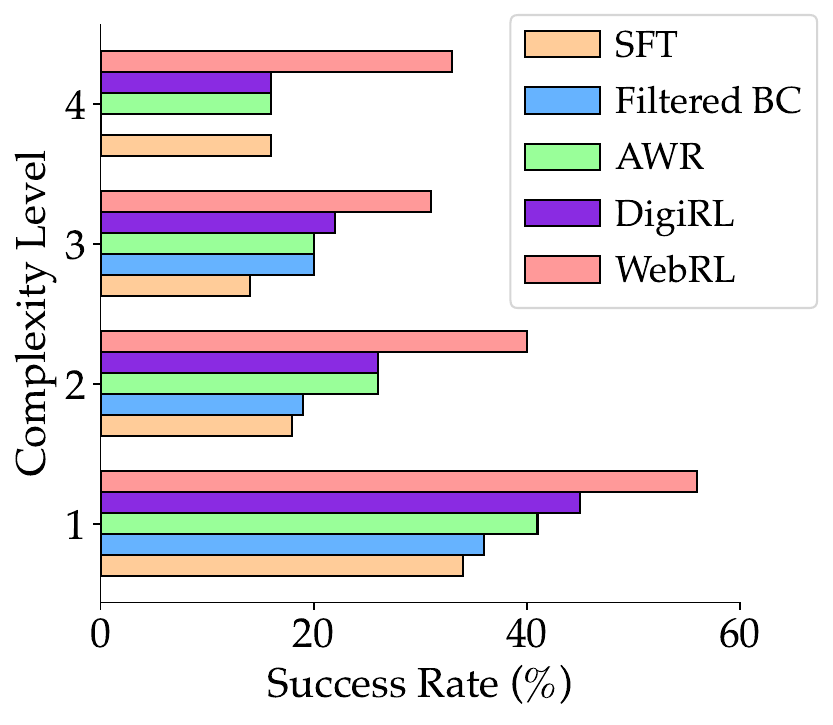}
    \caption{Accuracy of \model and baselines for tasks with different complexity.}
    \label{fig:complexity_analysis}
\end{wrapfigure}

We further analyze the performance of \model and baselines across instructions of varying complexity, as shown in Figure~\ref{fig:complexity_analysis}. Instruction complexity is measured by the number of requirements in the task. For example, the instruction ``What are the top-3 best-selling products in Jan 2023'' has two requirements: identifying the top-3 products and specifying the timeframe, giving it a complexity level of 2.
Our results show that \model performs well across different complexity levels, particularly excelling in more complex instructions. In contrast, while DigiRL uses online learning, it struggles with higher complexity due to its focus on a predefined set of tasks that do not align with the model's capabilities, limiting its adaptability. This highlights the effectiveness of our self-evolving curriculum learning strategy, which progressively increases task complexity based on the model’s capacity, enabling better performance on challenging tasks.

\subsection{Ablation Study}

We conduct an ablation study to evaluate the impact of the replay buffer, KL-constrained policy update algorithm, and the curriculum learning strategy on \model. To assess their contributions, we compare \model with four alternative models: (1) \model w/o replay buffer, where training uses only the current interaction trajectory, (2) \model w/o KL, where the policy is updated using REINFORCE with value function baseline (the gradient is $\mathbb{E}_{\nu} [(A(s, a, I) \nabla_\theta \log\pi_\theta(a|s, I) ]$) but retains the replay buffer, (3) \model w/o KL \& replay buffer, which uses neither a replay buffer nor the KL-constrained policy update algorithm, and (4) \model w/o CL, which ablates the curriculum learning approach, utilizing only the instructions generated in the first phase.

The results, shown in Figure~\ref{fig:ablation}, demonstrate that all the components used by \model are essential. \textbf{(1) The role of the replay buffer.} The results reveal that when the replay buffer is removed, both \model w/o replay buffer and \model w/o KL \& replay buffer experience worsening performance over time. This decline occurs because the models lose access to earlier experiences and focus only on recent data, leading to knowledge degradation.
\textbf{(2) The role of the KL-constrained policy update algorithm.} Comparing \model and \model w/o KL, \model consistently performs better, due to the incorporation of KL-constrained policy update algorithm. 
When the replay buffer is not used, the KL-constrained policy update algorithm degrades more slowly than REINFORCE with a value function baseline because it better retains past knowledge by controlling KL divergence. In contrast, REINFORCE with a value function baseline quickly overfits the current phase's data and consistently underperforms its initial value. Overall, the KL-constrained policy update algorithm is more effective at balancing the retention of past knowledge with the learning of new information.
\textbf{(3) The role of the self-evolving curriculum learning strategy.} When comparing \model to \model w/o CL, both exhibit an overall upward trend due to online learning. However, \model w/o CL progresses more slowly and reaches a lower performance ceiling because it operates within a fixed task framework, whereas \model generates new tasks that adapt to its evolving capabilities. This highlights the effectiveness of our self-evolving curriculum learning approach. Additional ablation studies to further investigate the effects of curriculum learning are presented in Figure~\ref{fig:ablation_w_digirl}.

\begin{wraptable}{r}{0.4\textwidth} 
    \centering
    \vspace{-3mm}
    \begin{threeparttable}
    \caption{The impact of perplexity in replay buffer filtering of \model.}
    \begin{tabularx}{0.4\textwidth}{>{\centering\arraybackslash}p{0.58cm} *{3}{>{\centering\arraybackslash}X}}
    \toprule
    {\makebox[0.5cm]{$\mathbf{[1, \infty]}$}} &
    {\makebox[0.5cm]{$\mathbf{[1, \frac{1}{0.95}]}$}} & 
    {\makebox[0.5cm]{$\mathbf{[\frac{1}{0.95},\! \frac{1}{0.5}]}$}} & 
    {\makebox[0.5cm]{$\mathbf{[\frac{1}{0.5},\! \infty]}$}} \\
    \midrule
    29.1 & 27.9 & 31.5 & 23.0 \\
    \bottomrule
\end{tabularx}
    \label{tab:perplexity}
    \end{threeparttable}
    \vspace{-3mm}
\end{wraptable}

\textbf{The influence of perplexity.} 
We analyze the impact of using perplexity to select data from the replay buffer for training. Various perplexity thresholds are tested in the first learning phase, and the results are summarized in Table~\ref{tab:perplexity}. It can be observed that training on data with very low perplexity (range [1, 1/0.95]) leads to performance deterioration. This suggests that repeatedly learning overly familiar data harms the model. Similarly, training exclusively on data with high perplexity (above 1/0.5) also degrades performance, likely due to the model struggling with unfamiliar data, causing a significant shift in policy distribution and hindering generalization. Optimal performance is achieved when training on data with a perplexity range of [1/0.95, 1/0.5], indicating that a balance between simple and complex data enhances model performance by focusing on moderately difficult examples.

\begin{wrapfigure}{r}{0.5\textwidth}  %
    \centering
    \vspace{-3mm}
    \includegraphics[width=1\linewidth]{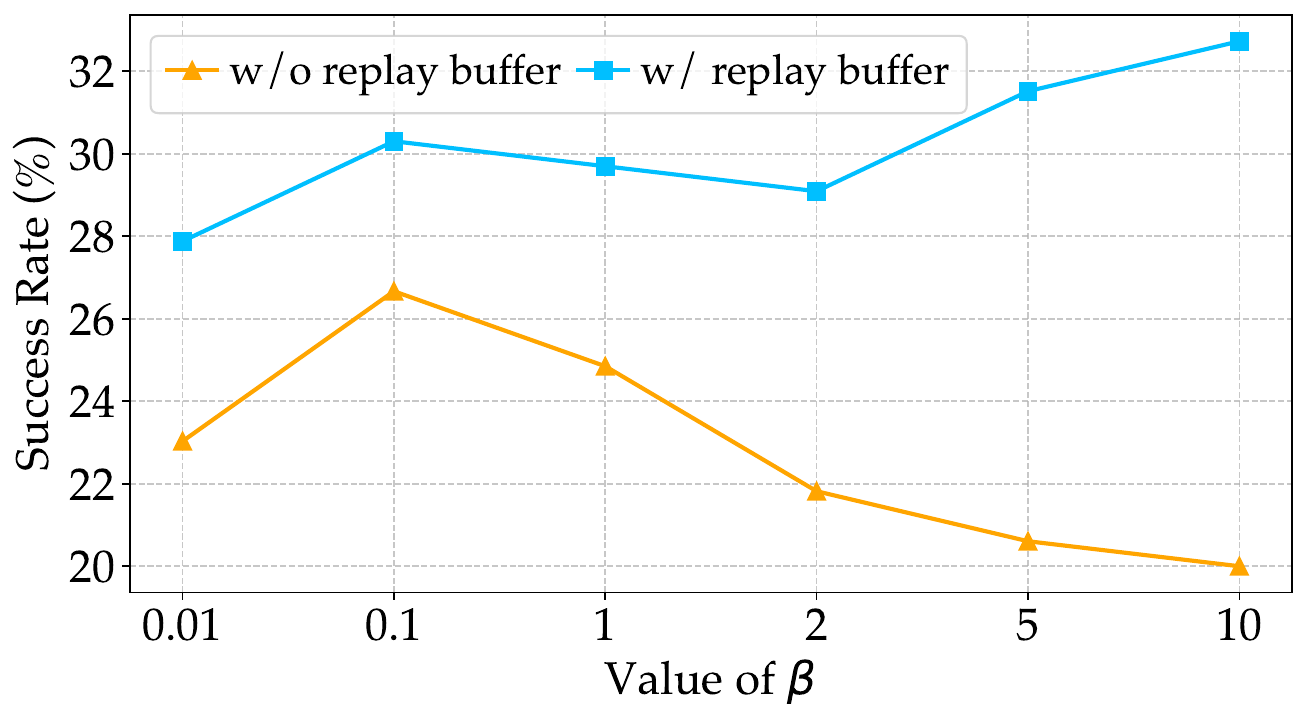}
    \caption{The impact of $\beta$ of KL-constrained policy update algorithm on the model's performance.}
    \label{fig:effect_of_beta}
    \vspace{-3mm}
\end{wrapfigure}

\textbf{The impact of $\beta$.} 
We investigate the effect of $\beta$ on performance with and without the replay buffer, as shown in Figure~\ref{fig:effect_of_beta}. The study is conducted on curriculum learning in one phase.
First, when $\beta$ is too small (e.g., $\beta$ = 0.01), model performance deteriorates, regardless of whether a replay buffer is used. This decline occurs because a small $\beta$ imposes a weak control over the KL divergence, causing the model to overfit the current data.
Second, without the replay buffer, performance initially improves as $\beta$ increases but then declines when $\beta$ becomes too large, indicating that large $\beta$ (e.g., $\beta \geq 1$) will overly restrict KL divergence, limiting the model's ability to update its policy effectively.
In contrast, with the replay buffer, performance stays high even at larger $\beta$ values. The historical experiences stored in the replay buffer facilitate more frequent and diverse parameter updates, supporting a stable improvement process, even as the $\beta$ value increases.

\input{table/orm}

\subsection{Evaluation of ORM}

In the \model framework, continuous improvement depends significantly on the effectiveness of the ORM, which plays a crucial role in evaluating interaction trajectories to guide the agent's learning process. To assess ORM's effectiveness, we compare its performance with several baseline models, including GPT-4-Turbo using identical inputs of our ORM, Captioner + GPT-4-Turbo, and GPT-4V, both using the same prompts with \citet{pan2024autonomous}.
We evaluate ORM and the baselines on two datasets: the WebArena-Lite test set and 100 sampled rollouts which are manually labeled. For the WebArena-Lite test data, we use its reward function outputs as labels. The results, shown in Table~\ref{tab:orm}, indicate that while the baseline models consistently achieve an accuracy slightly above 70\%, our ORM surpasses them with an accuracy of approximately 80\%.

\subsection{Case Study}

\begin{figure}[ht]
    \centering
    \includegraphics[width=0.95\linewidth]{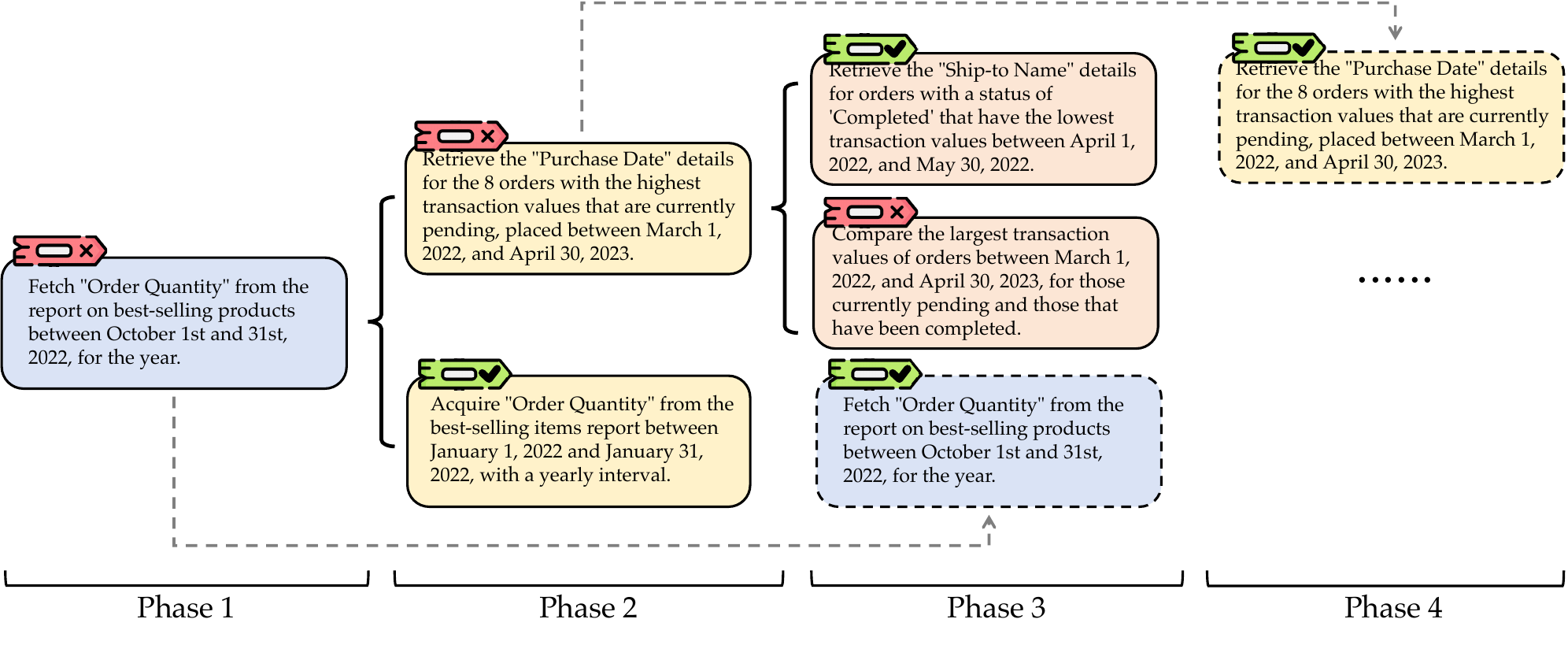}
    \caption{Examples of instructions generated in different phases under self-evolving curriculum learning.}
    \label{fig:instruction_case}
\end{figure}

Figure~\ref{fig:instruction_case} presents some instructions generated by the self-evolving curriculum learning strategy across different phases. Although these instructions are grouped by phase, the instructions shown in phase $i+1$ are not necessarily generated using the instructions from phase $i$ as seeds. 
As the phase increases, two types of augmentation occur for previously incomplete instructions. In one case, new instructions with similar task requirements or lower difficulty levels are generated. These new instructions help the model to successfully complete previously unfinishable tasks by allowing it to practice and build competence with the easier or related tasks first. For example, 
in phase 2, the instruction improves upon the phase 1 instruction by offering a clearer task description and explicitly requiring results in ``yearly interval". This enhancement enables the model to complete the task successfully by removing ambiguity. Additionally, changing the description of the instruction can improve agent exploration.
With the positive feedback from the clarified phase 2 instruction, the model can better understand and accurately perform the original phase 1 task as well.
In another case, tasks with increased complexity and diversity are generated. This task complication facilitates a continuous improvement in the model's capabilities by challenging its performance boundaries.
Therefore, the process of instruction generation exhibits such a pattern: for tasks that the model is unable to perform, analogous tasks are created to provide incremental steps that facilitate learning how to accomplish that type of task. Furthermore, tasks that remain challenging for the model are also generated and undergo the aforementioned iterative process. Through this cycle of challenge and refinement, the model’s capabilities gradually expand, enabling it to handle increasingly complex tasks over time.

%% file: table/main_result.tex
\begin{table*}[t]
\footnotesize
\setlength{\tabcolsep}{1pt} 
\centering
\begin{threeparttable}
\caption{Task success rate (SR) of \model and other comparison methods, evaluated on WebArena-Lite~\citep{zhou2023webarena,liu2024visualagentbench}, a human-verified subset of WebArena (* denotes results on full WebArena taken from literature reporting). The \textbf{best} and \underline{second-best} models are highlighted.}
\renewcommand\tabcolsep{1.5pt}
\renewcommand\arraystretch{1.02}

\begin{tabularx}{\textwidth}{p{5.4cm}*{6}{>{\centering\arraybackslash}X}p{1.1cm}<{\centering}}
\toprule
\textbf{Models} & \textbf{\#Params} & \textbf{Reddit} & \textbf{Gitlab} & \textbf{CMS} & \textbf{Map} & \textbf{OSS} & \textbf{Avg. SR}\\
\midrule
\multicolumn{8}{c}{\textit{Proprietary LLMs}} \\
\midrule
GPT-4-Turbo & N/A & 10.5 & 16.7 & 14.3 & 36.7 & 13.3 & 17.6 \\
GPT-4o & N/A & 10.5 & 10.0 & 20.0 & 20.0 & 11.1 & 13.9 \\
AWM + GPT-4-0613$^*$~\citep{wang2024agent} & N/A & 50.9 & 31.8 & 29.1 & \textbf{43.3} & 30.8 & 35.5 \\
WebPilot + GPT-4o$^*$~\citep{zhang2024webpilot} & N/A & \underline{65.1} & 39.4 & 24.7 & 33.9 & 36.9 & 37.2 \\
\midrule
\multicolumn{8}{c}{\textit{Open-sourced LLMs}} \\
\midrule
AutoWebGLM~\citep{lai2024autowebglm} & 6B & 9.4 & 15.0 & 28.6 & 24.8 & 17.1 & 18.2 \\
GLM-4-Chat~\citep{glm2024chatglm} & 9B & 5.3 & 10.0 & 6.7 & 3.3 & 6.7 & 6.1 \\
GLM-4 + SFT (BC) & 9B & 47.4 & 13.3 & 31.4 & 23.3 & 13.3 & 22.4 \\
GLM-4 + Filtered BC & 9B & 52.6 & 10.0 & 31.4 & 26.7 & 20.0 & 24.8 \\
GLM-4 + AWR~\citep{peng2019advantage} & 9B & 52.6 & 16.7 & 34.3 & 30.0 & 22.2 & 27.9 \\
GLM-4 + DigiRL~\citep{bai2024digirl} & 9B & 63.2 & 30.0 & 34.3 & 26.7 & 26.7 & 31.5 \\
GLM-4 + \model (ours) & 9B & 57.9 & \textbf{50.0} & \underline{48.6} & 36.7 & \underline{37.8} & \underline{43.0} \\
\midrule
Llama3.1-Instruct~\citep{dubey2024llama} & 8B & 0.0 & 3.3 & 2.9 & 3.3 & 11.1 & 4.8\\
Llama3.1 + SFT (BC) & 8B & 36.8 & 6.7 & 20.0 & 33.3 & 17.8 & 20.6 \\
Llama3.1 + Filtered BC & 8B & 52.6 & 20.0 & 31.4 & 23.3 & 8.9 & 23.0 \\
Llama3.1 + AWR~\citep{peng2019advantage} & 8B & 57.9 & 26.7 & 31.4 & 26.7 & 17.8 & 28.5 \\
Llama3.1 + DigiRL~\citep{bai2024digirl} & 8B & 57.9 & 26.7 & 37.1 & 33.3 & 17.8 & 30.3 \\
Llama3.1 + \model (ours) & 8B & 63.2 & \underline{46.7} & \textbf{54.3} & 36.7 & 31.1 & 42.4 \\ 
\midrule
Llama3.1-Instruct~\citep{dubey2024llama} & 70B & 10.5 & 16.7 & 17.1 & 20.0 & 4.4 & 12.7 \\
Llama3.1 + SFT (BC) & 70B & 52.6 & 20.0 & 20.0 & 26.7 & 13.3 & 23.0 \\
Llama3.1 + \model (ours) & 70B & \textbf{78.9} & \textbf{50.0} & \textbf{54.3} & \underline{40.0} & \textbf{44.4} & \textbf{49.1} \\
\bottomrule
\end{tabularx}
\label{tab:main_result}
\end{threeparttable}
\end{table*}

%% file: table/orm.tex
\begin{table}[t]
\footnotesize
\setlength{\tabcolsep}{1pt} 
\centering
\begin{threeparttable}
\caption{Evaluation on output-supervised methods (baselines adopted from~\citep{pan2024autonomous}). Our ORM, without accessing proprietary GPT-4, performs the best among all.}
\label{tab:orm}
\begin{tabular}{l>{\centering\arraybackslash}p{2.5cm}>{\centering\arraybackslash}p{2.5cm}>{\centering\arraybackslash}p{2.8cm}>{\centering\arraybackslash}p{2.5cm}}
\toprule
 & \textbf{Our ORM (8B)} & \textbf{GPT-4} & \textbf{Captioner + GPT-4} & \textbf{GPT-4V} \\
\midrule
\textbf{Test Dataset (\%)} & 80.8 & 71.9 & 72.6 & 71.2 \\ 
\textbf{Rollout (\%)}   & 79.4 & 71.2 & 73.3 & 70.5 \\ 
\bottomrule
\end{tabular}
\end{threeparttable}
\end{table}

%% file: section/related_work.tex
\section{Related Works}
\vpara{Adopting LLMs as Agent.}
As LLM capabilities advance, their applications extend beyond text generation~\citep{zheng2023judging, zhao2023survey} and complex reasoning~\citep{zelikman2024quiet, zhang2024codeagent}, and increasingly involve acting as agents for device control. Current research in this area falls into two main categories: training-free and training-based approaches. 
Training-free methods enhance pre-existing LLMs through prompt engineering~\citep{yan2023gpt, he2024webvoyager, zhang2024android} and constructing complex systems~\citep{liu2023bolaa, yang2023appagent, wang2023voyager, wu2024copilot, iong2024openwebagent, zhang2024ufo}. However, their performance is constrained by the limitations of the underlying LLMs, and the absence of fine-tuning restricts further improvement~\citep{chen2023fireact, zeng2023agenttuning, xie2024osworld}. 
Training-based approaches, primarily relying on imitation learning, require extensive expert demonstrations~\citep{zhang2023you, gur2023real, deng2024mind2web, hong2024cogagent, rawles2024androidinthewild, zhang2024agentohana}, which are costly to obtain. Although some methods use powerful LLMs like GPT-4 to generate demonstrations~\citep{chen2023fireact}, their accuracy remains insufficient for complex tasks. These methods often maximize the likelihood of individual actions without adequately considering long-term effects, limiting generalization~\citep{ghosh2021generalization, bai2024digirl}. 
To mitigate this, some studies use sampling-based methods to estimate long-term effects~\citep{lai2024autowebglm, putta2024agent}, while others, like ours, leverage reinforcement learning~\citep{carta2023grounding, bai2024digirl, pan2024autonomous, tantrue, zhai2024fine}. However, most existing methods rely on static task sets, which hinder the agent's continuous improvement as its capabilities evolve. To overcome this, we propose a dynamic task generation framework that adjusts task complexity based on the agent's progress, alongside a policy-update algorithm for ongoing performance enhancement.

\vpara{Reinforcement Learning for LLMs.}
Reinforcement learning (RL) has gained traction in training large language models (LLMs), with applications ranging from preference optimization~\citep{ouyang2022training, casper2023open} to complex reasoning~\citep{hao2023reasoning, pang2024iterative}. A growing area of interest involves using RL for device control tasks, which require multi-step interactions where the model selects appropriate actions based on the device state. This sequential decision-making aligns well with RL techniques. 
Existing research has explored RL-trained LLM agents for complex device control, primarily using online learning methods. For instance, AgentQ~\citep{putta2024agent} uses DPO~\citep{rafailov2024direct} for policy updates based on interaction data, while other methods~\citep{bai2024digirl, zhou2024archer, zhai2024fine} utilize actor-critic architectures, which we also adopt. However, in web tasks, feedback is often limited to binary success or failure after multiple interaction rounds. This can penalize correct intermediate actions due to later mistakes, complicating the reuse of previous data. 
Moreover, current research tends to focus on a fixed set of tasks for comparison with imitation learning, limiting the potential for continuous improvement through trial and error. To address this, we propose an autonomous curriculum learning mechanism that dynamically generates tasks based on the agent’s evolving skills, fostering ongoing progress. Additionally, we introduce a KL-constrained policy update algorithm and a specialized replay buffer to reuse valuable historical data and prevent knowledge forgetting during iterative curriculum updates.

%% file: section/conclusion.tex
\section{Conclusion}

In this work, we introduce \model, a novel self-evolving online curriculum reinforcement learning framework for training LLM-based web agents. By addressing key challenges including the scarcity of training tasks, feedback signal sparsity, and policy distribution drift, \model enables continual and consistent improvement in agent performance within online environments like WebArena. Our approach demonstrates substantial performance gains, significantly surpassing existing state-of-the-art web agents and proprietary LLM APIs. These results highlight the effectiveness of \model in advancing the capabilities of open-source LLMs for web-based tasks.

%% file: section/appendix/algorithm.tex
\section{Details of Policy Update Algorithm in \model}
\label{sec:algorithm}
\subsection{Derivation}
\label{sec:derivation}

By substituting eq.~\ref{eq: Q_V_relation} into eq.~\ref{eq: optimal_pi}, we can obtain:

\begin{equation} \label{eq:loss1}
    \beta \log \frac{\pi^*(a_t|s_t, I)}{\pi_{\text{ref}}(a_t|s_t, I)} = r(s_t, a_t, I) + V^*(s_{t+1}, I) - V^*(s_t) = A^*(s_t, a_t, I)
\end{equation}

\begin{equation} \label{eq:optimal_pi1}
    \pi^*(a_t|s_t, I) = \pi_{\text{ref}}(a_t|s_t, I) \exp \left(\frac{1}{\beta} A^*(s_t, a_t, I) \right)
\end{equation}

\(\pi^*\) is the optimal solution to our target (eq.~\ref{eq: target}). If \(\pi_\theta\) is represented by a function approximator, we aim to make it as close as possible to \(\pi^*\). 

\begin{equation} \label{eq:loss_func_1}
\begin{split}
    &\argmin\limits_{\theta} \E_{\nu} \left[ \left( \log\pi_\theta(a|s, I) - \log\pi^*(a|s, I) \right)^2 \right] \\
    = &\argmin\limits_{\theta} \E_{\nu} \left[ \left( \log\pi_\theta(a|s, I) - \log\pi_\text{ref}(a|s, I) - \frac{1}{\beta} A^*(s, a, I) \right)^2 \right] \\
    = &\argmin\limits_{\theta} \E_{\nu} \left[ \left( \beta \log \frac{\pi_\theta(a|s, I)}{\pi_{\text{ref}}(a|s, I)} - A^*(s, a, I) \right)^2 \right]
\end{split}
\end{equation}

Hence, the loss function can be defined as:

\begin{equation} \label{eq:final_loss}
    \mathcal{L}(\pi_\theta) = \E_{\nu} \left[ \left( \beta \log \frac{\pi_\theta(a|s, I)}{\pi_{\text{ref}}(a|s, I)} - A^*(s, a, I) \right)^2 \right]
\end{equation}

where $\nu$ represents the distribution of experience used for training. 
For any given state-action pair, \(\pi_\theta\) is expected to match the target policy \(\pi^*(a|s, I)\). There is no restriction on the data distribution \(\nu\), indicating the algorithm can function effectively in an off-policy setting. Since $A*$ is unknown, we estimate $A^*(s, a, I)$ using eq.~\ref{eq:loss_v} and eq.~\ref{eq:gae} to train the policy $\pi_\theta$. The feasibility of this approach is demonstrated in Appendix~\ref{sec:critic}.

\textbf{Further Analysis.}
Although \(\nu\) can follow any distribution, achieving stable policy improvement often requires some control over \(\nu\). The primary goal is to enhance the probability of actions that successfully complete the task and to address the deficiencies of the current policy $\pi_\theta$. Therefore, \(\nu\) typically consists of data sampled from the current policy being trained and prior successful experiences. 

\textbf{Why not use KL divergence to measure the distance between $\pi^*$ and $\pi_\theta$.}
For eq.~\ref{eq:loss_func_1}, many studies use KL divergence to measure the distance between two policies. When KL divergence is used, the optimization goal of $\theta$ is:

\begin{equation} \label{eq:loss_func_2}
\begin{split}
    &\argmin\limits_{\theta} \E_{s\sim d(s)} \left[ D_\text{KL}\left(\pi^*(\cdot | s, I) || \pi_\theta(\cdot|s, I)\right) \right] \\
    = &\argmax\limits_{\theta} \E_{s\sim d(s)} \E_{a\sim \pi^*(a|s, I)} \left[ \log \pi_\theta(a|s,I) \right] \\
    = &\argmax\limits_{\theta} \E_{s\sim d(s)} \int_{a} \pi_{\text{ref}}(a|s, I) \exp (\frac{1}{\beta} A^*(s, a, I) \log\pi_\theta(a|s,I) \text{d}a \\
    = &\argmax\limits_{\theta} \E_{s\sim d(s)} \E_{a \sim \pi_\text{ref}(a|s, I)} \left[ \log \pi_\theta(a|s, I) \exp(\frac{1}{\beta} A^*(s, a, I)) \right] 
\end{split}
\end{equation}
where $d(s)$ is a distribution of state. 
The choice of using eq.~\ref{eq:loss_func_1} instead of eq.~\ref{eq:loss_func_2} as the optimization target can be explained as follows:
(1) Eq.~\ref{eq:loss_func_2} imposes stronger restrictions on the training data. Specifically, it requires the training data to conform to the distribution of $\pi_\text{ref}$. 
\begin{wrapfigure}{r}{0.35\textwidth}  %
    \centering
    \includegraphics[width=1\linewidth]{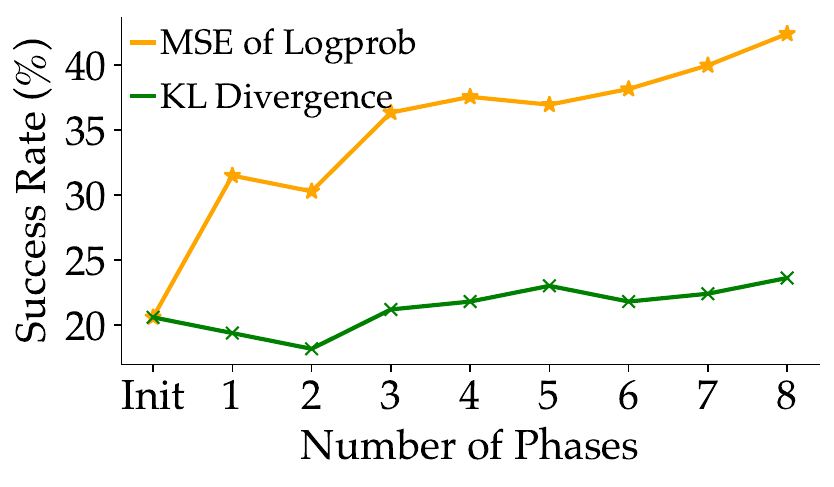}
    \vspace{-5mm}
    \caption{Comparison of  using eq.~\ref{eq:loss_func_1} and eq.~\ref{eq:loss_func_2} as target.}
    \label{fig:loss_comparison}
    \vspace{-3mm}
\end{wrapfigure}
When $\pi_\text{ref}$ fails to capture previous successful experiences, it becomes difficult to sample these experiences from $\pi_\text{ref}$, which in turn, can lead to further forgetting of those experiences. 
(2) Both eq.~\ref{eq:loss_func_1} and eq.~\ref{eq:loss_func_2} aim to approximate the distance between two distributions. The use of Mean Squared Error as a metric to measure the distance between policy distributions is also employed in \citet{fujimoto2021minimalist}. (3) Eq.~\ref{eq:loss_func_2} is unable to decrease the probability of actions with negative advantage. When correct action is hard to sample from \(\pi_{\text{ref}}\), eq.~\ref{eq:loss_func_2} may instead increase the probability of actions with negative advantage, compounding the challenge of sampling correct actions. 
We also conduct experiments to further validate the superiority of eq.~\ref{eq:loss_func_1}, with the results presented in Figure~\ref{fig:loss_comparison}. In these experiments, eq.~\ref{eq:loss_func_2} is trained exclusively on data sampled from \( \pi_{\text{ref}} \), while eq.~\ref{eq:loss_func_1} utilize both replay buffer data and sampled data.

\textbf{When discount factor $\gamma$ is used.} When considering the discount factor,  eq.~\ref{eq:loss1} is modified as follows:
\begin{equation}
    \beta \log \frac{\pi^*(a_t|s_t, I)}{\pi_{\text{ref}}(a_t|s_t, I)} = r(s_t, a_t, I) + \gamma V^*(s_{t+1}, I) - V^*(s_t) = A^*(s_t, a_t, I)
\end{equation}
This will not affect the subsequent derivation of the policy loss function (eq.~\ref{eq:final_loss}). The loss function for the critic remains unchanged. However, the equation to calculate the advantage is modified as follows:
\begin{equation}
    A(s_t,\!a_t,\!I)\!=\!\lambda \big(r(s_t,\!a_t,\!I) + \gamma V(s_{t+1},\!I)\!-\!V(s_t,\!I)\big)\!+\!(1\!-\!\lambda) \big(\gamma^{T-t} r(s_T,\!a_T,\!I)\!-\!V(s_t,\!I)\big)
\end{equation}
In our experiment, we use the discount factor and set its value to 0.9.

\
We use eq.~\ref{eq:loss_v} and eq.~\ref{eq:gae} to estimate $A^*(s, a, I)$. The training of critic (eq.~\ref{eq:loss_v}) also operates in an off-policy manner.

\subsection{Proof}
\label{sec:critic}

To prove the feasibility of replacing \( A^* \) with an advantage estimator \( A \) based on a value network \( V \), which is trained on data collected throughout the training process, we aim to demonstrate that using \( V \) to train the policy can lead to policy improvement.

The current policy is denoted by $\pi_\text{old}$. The training dataset $D$ is composed of trajectories sampled using $\pi_\text{old}$ as well as historical experiences stored in a replay buffer. The behavioral policy associated with the dataset $D$ is referred to as $\pi_{\mu}$. The corresponding value function and action-value function under $\pi_{\mu}$ are denoted as $V^{\pi_{\mu}}$ and $Q^{\pi_{\mu}}$,  respectively. Below, we demonstrate that minimizing the objective $\arg\min_\theta \E_D[(\log_{\pi_\theta}(a_t|s_t) - \exp (\frac{Q^{\pi_\mu}(s_t, a_t) - V^{\pi_\mu}(s_t)}{\beta}))^2]$ can lead to policy improvement. For brevity, we omit $I$ in the following proof.

Let the new policy $\pi_\text{new}(a_t|s_t)$ be defined as:
\begin{equation}
    \pi_\text{new}(a_t|s_t) = \exp\left(\frac{Q^{\pi_\mu}(s_t, a_t) - V^{\pi_\mu}(s_t)}{\beta}\right)
\end{equation}
It can be shown that:  $D_\text{KL}(\pi_\text{new} ||\pi_\text{new})
\leq D_\text{KL}(\pi_\mu ||\pi_\text{new})$. From this, we derive the following inequality:
\begin{equation}
\begin{split}
    &\mathbb{E}_{a_t \sim \pi_\text{new}}\!\left[\log \pi_\text{new}(a_t|s_t)\!-\!\frac{Q^{\pi_\mu}(s_t, a_t)}{\beta}\!+\!\frac{V^{\pi_\mu}(s_t)}{\beta}\right]
\\
\leq &\ \mathbb{E}_{a_t \sim \pi_\mu}\!\left[\log \pi_\mu(a_t|s_t)\!-\!\frac{Q^{\pi_\mu}(s_t, a_t)}{\beta}\!+\!\frac{V^{\pi_\mu}(s_t)}{\beta} \right]
\end{split}
\end{equation}
Since $V^{\pi_\mu}(s_t)$ depends only on the state \(s_t\) and not on the actions, it can be factored out of the expectation. We can rewrite the inequality as:
\begin{equation}\label{eq:bound}
    \mathbb{E}_{a_t \sim \pi_\text{new}} \left[ Q^{\pi_\mu}(s_t, a_t) - \beta \log \pi_\text{new}(a_t|s_t) \right] \geq V^{\pi_\mu}(s_t)
\end{equation}
where $V^{\pi_\mu}(s_t) = \mathbb{E}_{a\sim \pi_\mu(a_t|s_t)} [Q^{\pi_\mu}(s_t, a_t) -\beta \log \pi_\mu (a_t|s_t)]$, which is satisfied in maximum entropy reinforcement learning setting~\citep{haarnoja2018soft}.

Consider the soft Bellman equation:
\begin{equation}
    \begin{split}
        Q^{\pi_\mu}(s_t, a_t) &= r(s_t|a_t) + \beta \log\pi_\text{ref}(a_t|s_t) + V^{\pi_\mu}(s_{t+1}) \\
        &\leq r(s_t|a_t) + \beta \log\pi_\text{ref}(a_t|s_t) + \mathbb{E}_{a_{t+1}\sim \pi_\text{new}}[Q^{\pi_\mu}(s_{t+1}, a_{t+1}) - \beta \log \pi_\text{new} (a_{t+1}|s_{t+1})] \\
        &\cdots \\
        &\leq Q^{\pi_\text{new}}(s_t, a_t)
    \end{split}
\end{equation}

where we iteratively expand $Q^{\pi_\mu}$ by applying the soft Bellman equation and the bound provided in Eq~\ref{eq:bound}.
Since $\pi_\text{new}$ is an improved policy over $\pi_\mu$ (i.e., $Q^{\pi_\text{new}(s_t, a_t)} \geq Q^{\pi_\mu}(s_t, a_t)$), and $\pi_\theta$ is trained to closely match $\pi_\text{new}$, it follows that $\pi_\theta$ will also be an improved policy over $\pi_\mu$ provided the approximation is accurate. Formally, we have: 
\begin{equation}
    V^{\pi_\theta}(s) \approx V^{\pi_\text{new}(s)} \geq V^{\pi_\mu}(s)
\end{equation}

Since we only incorporate the correct trajectory from the historical data, we can assert that policy \(\pi_\mu\) will not perform worse than the old policy \(\pi_\text{old}\). Therefore, we have the relationship:

\begin{equation}
    V^{\pi_\theta}(s) \approx V^{\pi_\text{new}}(s) \geq V^{\pi_\mu}(s) \geq V^{\pi_\text{old}}(s)
\end{equation}
The policy improvement process is well-established. By iteratively performing policy evaluation and policy improvement, the policy \(\pi_\theta\) can be made to converge toward the optimal policy \(\pi^*\). 
If the data in the replay buffer is not utilized, the same theoretical process holds. However, in this case, $V^{\pi_\mu}$ and $Q^{\pi_\mu}$ are replaced by \(V^{\pi_\text{old}}\) and \(Q^{\pi_\text{old}}\).

\begin{wrapfigure}{r}{0.35\textwidth}  %
    \centering
    \vspace{-6mm}
    \includegraphics[width=1\linewidth]{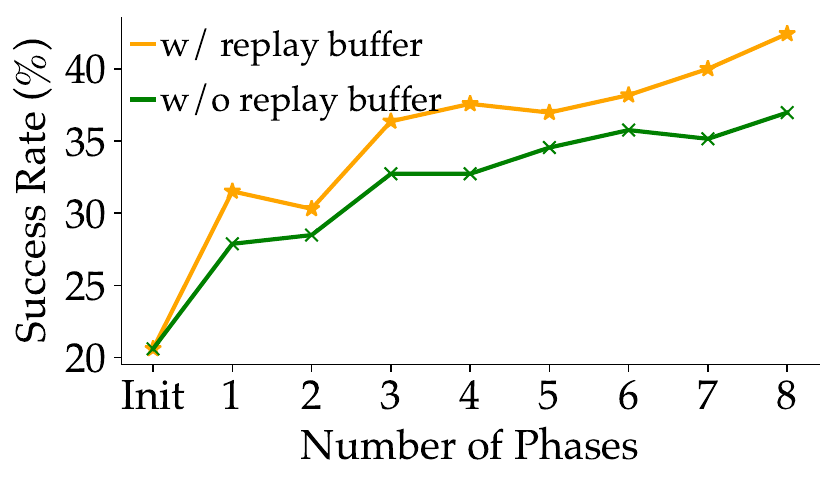}
    \vspace{-5mm}
    \caption{Comparison of  using $V^{\pi_\mu}$ and $V^{\pi_\text{old}}$.}
    \label{fig:critic}
    \vspace{-3mm}
\end{wrapfigure}

We conduct an experiment to evaluate the effectiveness of using \( V^{\pi_\mu} \) instead of \( V^{\pi_\text{old}} \) in \model training. In this experiment, the policy \( \pi_\theta \) is trained using data from two sources: rollout trajectories during the current phase and past experiences stored in the replay buffer. For the critic, we compare two training settings: training with data from both rollout trajectories and experiences in the replay buffer versus using only rollout trajectories.
The results, as presented in Figure~\ref{fig:critic}, reveal that the model performs better when the critic is trained with data from both the replay buffer and the rollout trajectories. This observation underscores the significance of integrating experiences from prior successful trajectories into the critic's training process, which can lead to more effective policy improvement.

\subsection{Comparison}

The comparison result between the policy update algorithm of \model and other algorithms is shown in Table~\ref{tab:comparison}. 

\input{table/appendix/comparison}

\textbf{Comparison with DPO~\citep{rafailov2024direct}. } DPO also solves for the optimal value of eq.~\ref{eq: target}, obtaining the relationship between the optimal policy \(\pi^*\) and the reward \(r\). Rather than explicitly solving for the reward \(r\), DPO employs contrastive learning by constructing paired data to build the learning target of \(\pi_\theta\). This approach allows DPO to bypass the need to estimate \(r\). In contrast, our approach explicitly fits the value function \(V\) to guide the optimization of \(\pi_\theta\), making it converge towards the optimal policy \(\pi^*\). \textbf{However, DPO requires pair-wise data, which is challenging to obtain in web tasks.} This is primarily because implementing state backtracking on web pages is difficult, making it hard to collect outcomes of different actions on the same page state.

\textbf{Comparison with PPO~\citep{schulman2017proximal}. }
Eq.\ref{eq: target} is also the target of RLHF. Prior work often employs PPO to optimize this objective. \textbf{However, as an on-policy algorithm, PPO has low sampling efficiency and requires large amounts of data for stable improvement}. This limitation makes it unsuitable for environments such as WebArena and real-world websites, where interaction is expensive and inefficient. Instead, our approach operates in an off-policy manner.

\textbf{Comparison with AWR~\citep{peng2019advantage}. }
AWR is one of the widely used offline RL algorithms. The loss function of AWR is $\argmax_\theta \E_\nu \left[ \log \pi_\theta (a|s, I) \cdot \exp (A(s, a, I) / \beta \right]$. \textbf{Compared to our method, AWR has a different optimization goal.} Our approach explicitly minimizes the KL divergence between the reference policy \(\pi_{\text{ref}}\) and the current policy \(\pi_{\theta}\), while AWR focuses on constraining the KL divergence between the current policy and the behavioral policy derived from training data $\nu$ (instead of the reference policy \(\pi_{\text{ref}}\)). Additionally, \textbf{AWR does not directly reduce the probability of incorrect actions.} In web tasks, sampled data often include unsuccessful traces. Ignoring such data wastes valuable information, while using it risks increasing the likelihood of incorrect actions. In contrast, our method effectively reduces the probability of incorrect actions.

%% file: table/appendix/comparison.tex
\begin{table}[h!] 
    \centering
    \caption{Comparison between \model and different algorithms based on various criteria.}
    \label{tab:comparison}
    \begin{tabular}{l>{\centering\arraybackslash}p{1.5cm}>{\centering\arraybackslash}p{2.2cm}>{\centering\arraybackslash}p{3.3cm}>{\centering\arraybackslash}p{2.8cm}}
        \toprule
        {\fontsize{9.5pt}{9pt}\selectfont \textbf{Algorithm}} & {\fontsize{9.5pt}{9pt}\selectfont \textbf{Off-policy}} & {\fontsize{9.5pt}{9pt}\selectfont \textbf{Pair-wise Data}} & {\fontsize{9.5pt}{9pt}\selectfont \textbf{KL-constrained Target}} & {\fontsize{9.5pt}{9pt}\selectfont \textbf{Reduce Error Prob}} \\
        \hline
        \model & \raisebox{-2.2pt}{\includegraphics[scale=0.22]{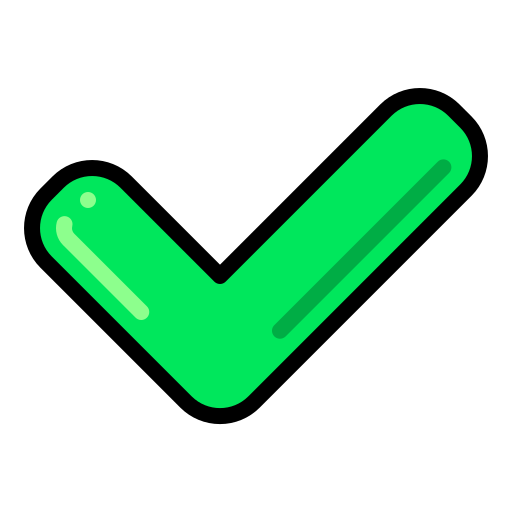}} & \raisebox{-1.8pt}{\includegraphics[scale=0.35]{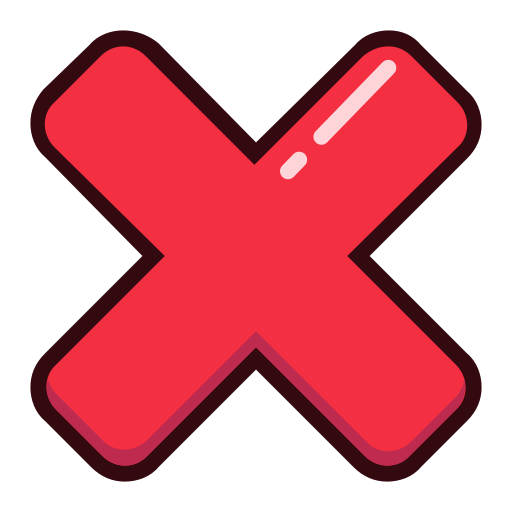}} & \raisebox{-2.2pt}{\includegraphics[scale=0.22]{figure/emoji/check.png}} & \raisebox{-2.2pt}{\includegraphics[scale=0.22]{figure/emoji/check.png}} \\
        DPO & \raisebox{-2.2pt}{\includegraphics[scale=0.22]{figure/emoji/check.png}} & \raisebox{-2.2pt}{\includegraphics[scale=0.22]{figure/emoji/check.png}} & \raisebox{-2.2pt}{\includegraphics[scale=0.22]{figure/emoji/check.png}} & \raisebox{-2.2pt}{\includegraphics[scale=0.22]{figure/emoji/check.png}} \\
        PPO & \raisebox{-1.8pt}{\includegraphics[scale=0.35]{figure/emoji/delete.png}} & \raisebox{-1.8pt}{\includegraphics[scale=0.35]{figure/emoji/delete.png}} &  \raisebox{-2.2pt}{\includegraphics[scale=0.22]{figure/emoji/check.png}} & \raisebox{-2.2pt}{\includegraphics[scale=0.22]{figure/emoji/check.png}} \\
        AWR & \raisebox{-2.2pt}{\includegraphics[scale=0.22]{figure/emoji/check.png}} & \raisebox{-1.8pt}{\includegraphics[scale=0.35]{figure/emoji/delete.png}} & \raisebox{-1.8pt}{\includegraphics[scale=0.35]{figure/emoji/delete.png}} & \raisebox{-1.8pt}{\includegraphics[scale=0.35]{figure/emoji/delete.png}}\\
        \bottomrule
    \end{tabular}
\end{table}

%% file: section/appendix/baselines.tex
\section{Training Details} 
\label{sec:training}

\subsection{Details of \model}

\textbf{The agent observation} consists of three components:
\begin{packeditemize}
    \item User instruction: The instruction provided by the user.
    \item Action history: A record of the actions the agent has previously taken.
    \item Webpage HTML: HTML content of the current webpage.
\end{packeditemize}
We process the HTML, simplifying its structure and assigning distinct element IDs to all clickable elements. This facilitates the model's ability to identify and indicate which specific element requires manipulation.

\begin{figure}[ht]
    \centering
    \includegraphics[width=\linewidth]{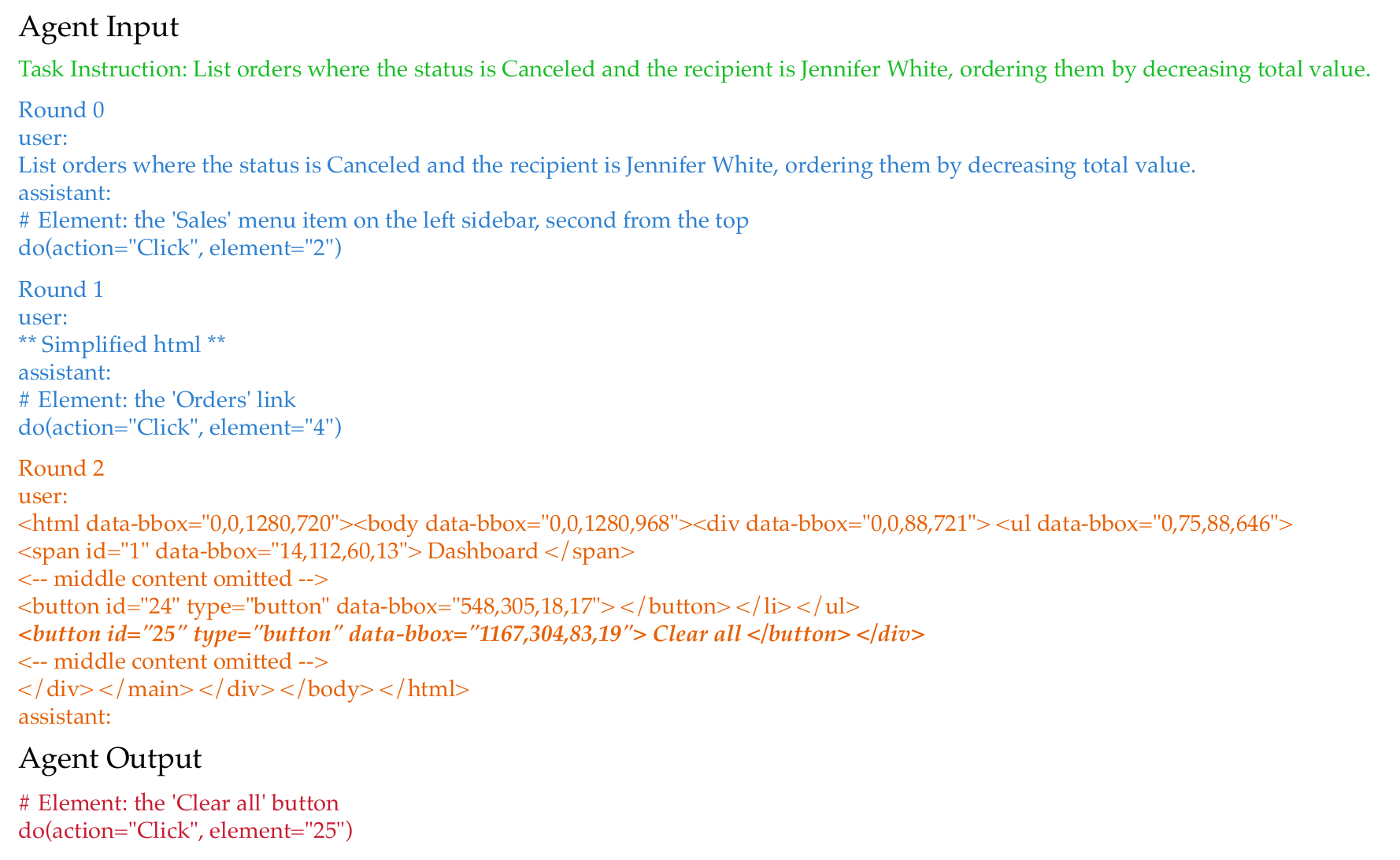}
    \caption{The input and output format of \model and baselines, where the input is composed of task instruction (in \textcolor{mygreen}{green}), action history (in \textcolor{myblue}{blue}), and HTML of the current webpage (in \textcolor{myorange}{orange}). The output (in \textcolor{myred}{red}) is the action taken on the current webpage.}
    \label{fig:prompt_for_agent}
\end{figure}

\textbf{The agent actions} are mostly similar to those defined in WebArena-Lite~\citep{liu2024visualagentbench}, with some additional actions. 
\begin{packeditemize}
    \item Click: Clicks an element with a specific ID.
    \item Hover: Hovers over an element with a specific ID.
    \item Type: Types a message into an input box with a specific ID.
    \item Search: Types a message into an input box with a specific ID and presses Enter to initiate a search.
    \item Press: Emulates a specific keyboard key combination.
    \item Scroll: Scrolls the page up or down.
    \item Select\_dropdown\_option: Selects an option from a dropdown menu with a specific ID.
    \item New\_tab: Opens a new tab in the current browser.
    \item Tab\_focus: Switches focus to a browser tab at a specified index.    \item Close\_tab: Closes the current tab.
    \item Goto: Navigates to a specific URL.
    \item Go\_back: Returns to the previous page.
    \item Go\_forward: Moves to the next page if available.
    \item Exit: Terminates the operation, returns the response, and exits.
\end{packeditemize}
To provide more detailed information about the action, we include comments labeled with ``\# Element:'' in the action, which describe the operated element. Similarly, we include comments labeled ``\# Note:'', which quote relevant information from the current webpage that supports completing the instruction. 
An example of the agent's specific input and output is shown in Figure~\ref{fig:prompt_for_agent}. The input is the observation, while the output specifies the action to be performed on the webpage. The ``element'' argument identifies the target element for the action. More examples can be found in Appendix~\ref{sec:examples}.

\textbf{The detailed training process of \model} is shown in Algorithm~\ref{algorithm:1}. First, we perform supervised fine-tuning using the WebArena-Lite training dataset. We then initialize the replay buffer and failure set by running the SFT-trained model on instructions corresponding to the WebArena-Lite training dataset. Subsequently, in each phase of the self-evolving curriculum reinforcement learning process, 500 new instructions that meet the filtering criteria are selected from those generated by GPT-4o. Both newly generated interaction data on these instructions and historical data with perplexity between 1/0.95 and 1/0.5 from the replay buffer are used to train the actor and critic. The amount of historical data used is limited to twice the size of the interaction data. The hyperparameters employed in \model are presented in Table~\ref{tab:hyperparameter}.

\subsection{Details of Baselines}

For RL-based baselines (except DigiRL), the interaction data from \model's first phase is used, while DigiRL is trained using the first-phase instructions in an online learning setup. Hence, except for DigiRL, the other RL baselines fall under offline reinforcement learning.

We reproduce the same framework used in DigiRL within the WebArena environment. Specifically, we use the same components, including the AWR method, the instruction-level and step-level value functions, and the replay buffer described in DigiRL. To apply DigiRL in WebArena-Lite, the main modifications we make are adjusting the data format to align with \model and tweaking certain hyperparameters. DigiRL also conducts 8 rounds of interaction and training. The hyperparameters employed in those baselines are presented in Table~\ref{tab:hyperparameter}.

\subsection{Details of ORM}

WebArena-Lite provides 1,186 training samples each comprising an instruction, a trajectory, and a reward function. To train the \orm{}, we first enhance the WebArena-Lite training dataset by rewriting the instructions. Then, we collect rollouts from all baseline methods on this augmented dataset. These rollouts, combined with the original WebArena-Lite training samples, form the complete training dataset for the \orm{}, with a total of 12,200 samples. We use the associated reward function to label these trajectories, identifying whether they successfully complete the task. These trajectories are subsequently used to train the ORM, with the specific hyperparameters listed in Table~\ref{tab:hyperparameter_orm}. The prompt for \orm{} is shown in Figure~\ref{fig:prompt_for_reward_model}. \orm{} is required to produce either ``YES" or ``NO'' as its output. To determine the evaluation result, we compare the probabilities assigned to ``YES'' and ``NO'' and select the one with the higher probability. We need to train \orm{} because new tasks that do not have predefined reward functions will be generated in the subsequent self-evolving curriculum reinforcement learning process. We rely on the trained \orm{} to judge whether each task is successfully completed.

\begin{algorithm} 
\caption{\model}
\label{algorithm:1}
\textbf{Input:} WebArena-Lite training set $D_0$ and corresponding instruction set $I_0$, Base model $\mathcal{M}_\text{base}$, Replay buffer $\mathcal{B}$, Failure set $\mathcal{F}$, Reward model $\mathcal{M}_\text{ORM}$, Number of phases $N$, Number of instructions per phase $K$

\vspace{0.18em}{\hspace{-0.5em}\fontsize{10pt}{10pt}\selectfont\textbf{\hspace{1em}Part 1. SFT Training}}

\begin{algorithmic}[1]
    \STATE $\mathcal{M}_\text{SFT}$ $\longleftarrow$ perform supervised fine-tuning on model $\mathcal{M}_\text{base}$ with dataset $D_0$
    \STATE $D_\text{rollout}$ $\longleftarrow$ rollout trajectories of $\mathcal{M}_\text{SFT}$ on $I_0$
    \STATE $D_\text{success}$, $D_\text{fail}$ $\longleftarrow$ evaluate $D_\text{rollout}$ using WebArena-Lite reward function
    \STATE $\mathcal{B}$ $\longleftarrow$ $D_0$ \textcolor{gray!80}{\# Initialize replay buffer with successful trajectories}
    \STATE $\mathcal{F}$ $\longleftarrow$ $D_\text{fail}$ \textcolor{gray!80}{\# Initialize failure set with instructions of failing trajectories}\\
    \vspace{0.18em}{\hspace{-1.2em}\fontsize{10pt}{10pt}\selectfont\textbf{Part 2. Self-evolving Curriculum RL}}\vspace{0.15em}
    \STATE $\pi_1$ $\longleftarrow$ $\mathcal{M}_\text{SFT}$ \textcolor{gray!80}{\# Initialize actor/policy}
    \STATE $V_1$ $\longleftarrow$ $\mathcal{M}_\text{SFT}$ with a randomly initialized value head \textcolor{gray!80}{\# Initialize critic}
    \FOR{$n$ in 1...$N$}
    \STATE $I_n$ $\longleftarrow$ \{\}
    \WHILE{size($I_n$) $\leq$ $K$}
    \STATE $I_\text{generation}$ $\longleftarrow$ instructions generated by GPT-4o with instructions from $\mathcal{F}$ as examples \textcolor{gray!80}{\# instruction generation process} 
    \STATE $I_\text{filter}$ $\longleftarrow$ filter($I_\text{generation}$) \textcolor{gray!80}{\# instruction filtering process} 
    \STATE $I_n$ $\longleftarrow$ $I_n \cup I_\text{filter}$
    \ENDWHILE
    \STATE $D_\text{rollout}$ $\longleftarrow$ rollout trajectories of $\pi_n$ on $I_n$
    \STATE $D_\text{success}$, $D_\text{fail}$ $\longleftarrow$ evaluate $D_\text{rollout}$ with $\mathcal{M}_\text{ORM}$ \textcolor{gray!80}{\# use $\mathcal{M}_\text{ORM}$ to label the rollout trajectories}
    \STATE $D_\text{experience}$ $\longleftarrow$ experiences from $\mathcal{B}$ with perplexity computed by $\pi_n$ between $\frac{1}{0.95}$ and $\frac{1}{0.5}$
    \STATE $\pi_{n+1}$,\! $V_{n+1}$ $\longleftarrow$ train($\pi_n$, $V_n$, $D_\text{rollout} \cup D_\text{experience}$) \textcolor{gray!80}{\# use loss functions from eq.~\ref{eq:loss} and eq.~\ref{eq:loss_v}}
    \STATE $\mathcal{B}$ $\longleftarrow$ $\mathcal{B} \cup D_\text{success}$
    \STATE  $\mathcal{F}$ $\longleftarrow$ $\mathcal{F} \cup D_\text{fail}$
    \ENDFOR
\end{algorithmic} 
\end{algorithm}

\input{table/appendix/hyperparameter}

\input{table/appendix/hyperparameter_orm}

%% file: table/appendix/hyperparameter.tex
\begin{table*}[ht]
\footnotesize
\setlength{\tabcolsep}{1pt} 
\centering
\begin{threeparttable}
\caption{The hyperparameters we employ in \model and baselines.}
\renewcommand\tabcolsep{1.5pt}
\begin{tabularx}{0.9\textwidth}{p{2.5cm}*{1}{>{\centering\arraybackslash}X}p{3cm}<{\centering}}
\toprule 
\textbf{Method} & \textbf{Hyperparameter} & \textbf{Value} \\
\midrule
\multirow{6}{*}{SFT} & learning rate & 1e-5 \\
& lr scheduler type & cosine \\
& warmup ratio & 0.1 \\
& batch size & 128 \\
& training epoch & 1 \\
& cutoff length & 16384 \\
\midrule
\multirow{6}{*}{Filtered BC} & learning rate & 1e-6 \\
& lr scheduler type & constant \\
& batch size & 128 \\
& training epoch & 1 \\
& cutoff length & 16384 \\
& filtering threshold & 70th percentile \\
\midrule
\multirow{8}{*}{AWR} & actor learning rate & 1e-6 \\
& actor lr scheduler type & constant \\
& critic learning rate & 1e-6 \\
& critic lr scheduler type & constant \\
& batch size & 128 \\
& discount factor & 0.9 \\
& actor training epoch & 1 \\
& critic training epoch & 1 \\
\midrule
\multirow{13}{*}{DigiRL} & actor learning rate & 1e-6 \\
& actor lr scheduler type & constant \\
& critic learning rate & 1e-6 \\
& critic lr scheduler type & constant \\
& instruction value function lr & 1e-6 \\
& instruction value function lr scheduler type & constant \\
& batch size & 128 \\
& discount factor & 0.9 \\
& actor training epoch & 1 \\
& critic training epoch & 1 \\
& instruction value function epoch & 1 \\
& rollout temperature & 1 \\
& replay buffer size & 100000 \\
\midrule
\multirow{9}{*}{\model} & actor learning rate & 1e-6 \\
& actor lr scheduler type & constant \\
& critic learning rate & 1e-6 \\
& critic lr scheduler type & constant \\
& batch size & 128 \\
& discount factor & 0.9 \\
& actor training epoch & 1 \\
& critic training epoch & 1 \\
& rollout temperature & 1 \\
\bottomrule
\end{tabularx}
\label{tab:hyperparameter}
\end{threeparttable}
\end{table*}

%% file: table/appendix/hyperparameter_orm.tex
\begin{table*}[ht]
\footnotesize
\setlength{\tabcolsep}{1pt} 
\centering
\begin{threeparttable}
\caption{The hyperparameters we employ to train the ORM.}
\renewcommand\tabcolsep{1.5pt}
\begin{tabularx}{0.6\textwidth}{>{\centering\arraybackslash}p{4.5cm}>{\centering\arraybackslash}X}
\toprule 
\textbf{Hyperparameter} & \textbf{Value} \\
\midrule
learning rate & 5e-6 \\
lr scheduler type & cosine \\
warmup ratio & 0.1 \\
batch size & 128 \\
training epoch & 4 \\
cutoff length & 16384 \\
\bottomrule
\end{tabularx}
\label{tab:hyperparameter_orm}
\end{threeparttable}
\end{table*}

%% file: section/appendix/more_results.tex
\section{Other Quantitative Experiments}

\begin{figure}[h]
    \centering
    \includegraphics[width=0.9\linewidth]{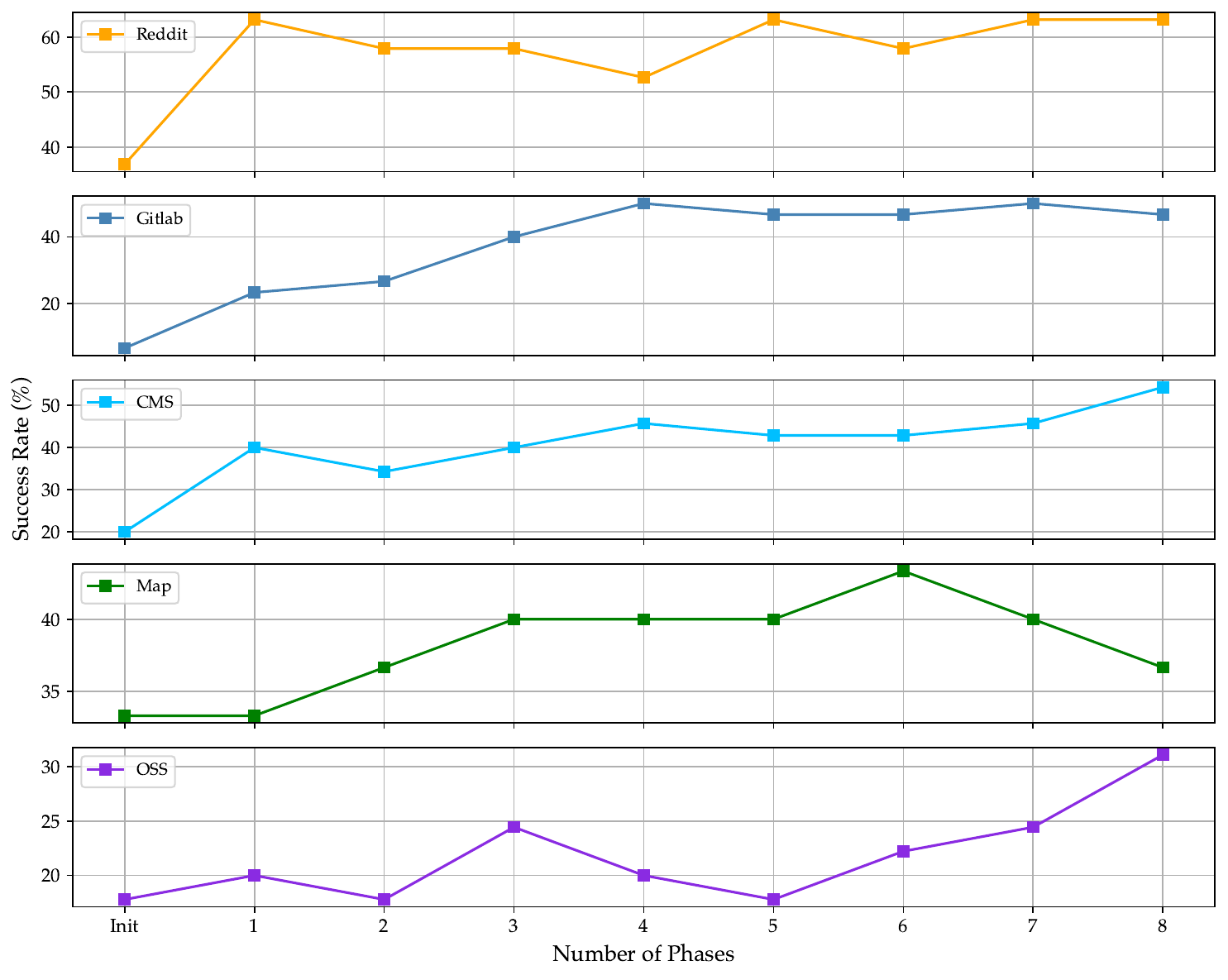}
    \caption{Performance variation curves of Llama3.1-8B on each website under WebRL training}
    \label{fig:curve_of_different_web}
\end{figure}

Figure~\ref{fig:curve_of_different_web} illustrates the performance variation curves of Llama3.1-8B trained with \model on each website. It can be seen that in all the sites except for Map, there is a clear upward trend. However, in the case of Map, there is an initial upward trend followed by a decline. We hypothesize that the final decline is caused by a significant increase in OSS and CMS implementation, which creates a trade-off. This trade-off leads to a performance drop in Map and a slight decline in GitLab.

\begin{figure}[t]
  \centering
  \begin{minipage}[t]{0.43\textwidth}
    \centering
    \includegraphics[width=1.0\linewidth]{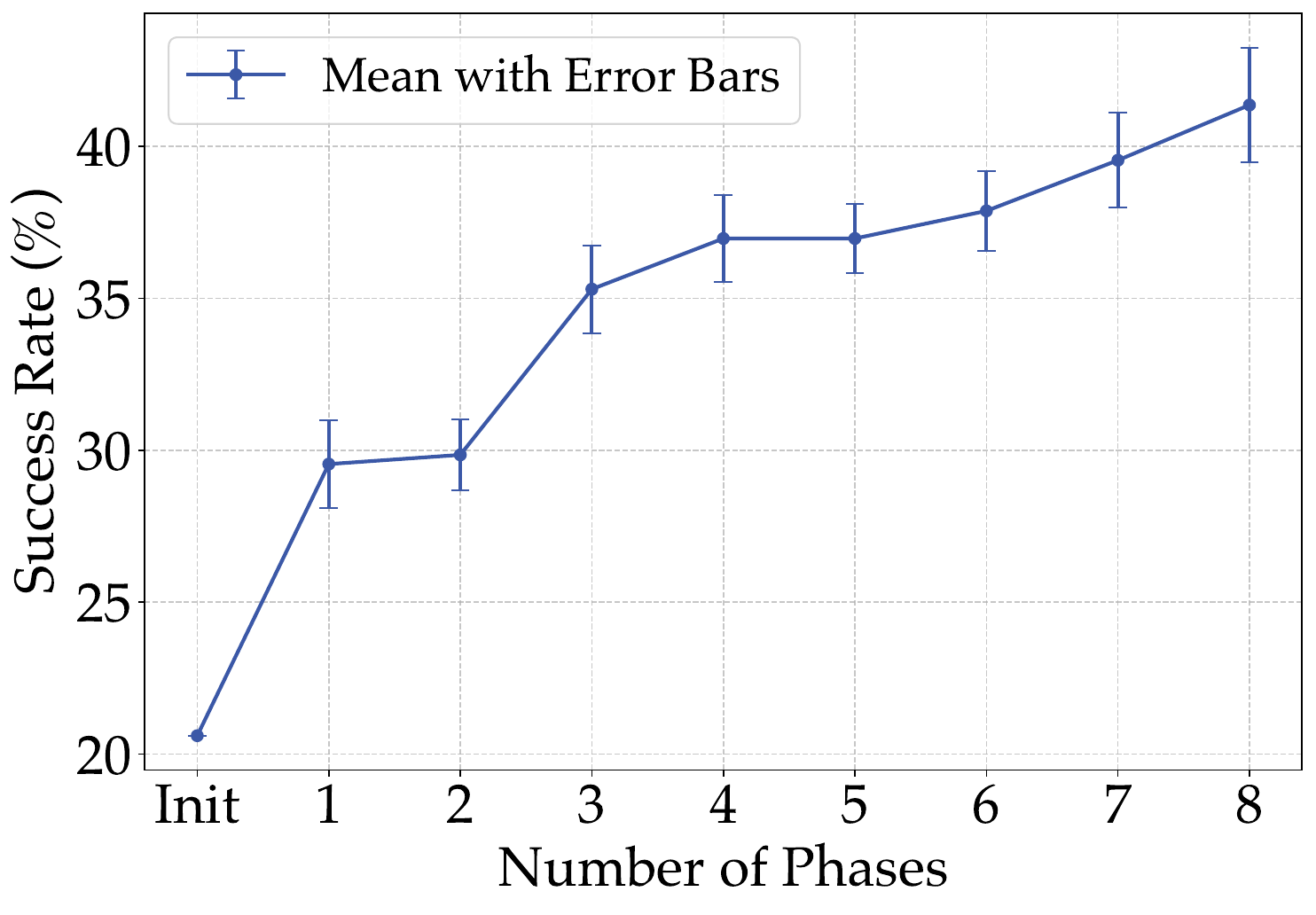}
    \vspace{-5mm}
    \caption{Error bars of \model  for each phase.}
    \label{fig:error_bars}
  \end{minipage}
  \hfill
  \begin{minipage}[t]{0.55\textwidth}
    \centering
    \includegraphics[width=\linewidth]{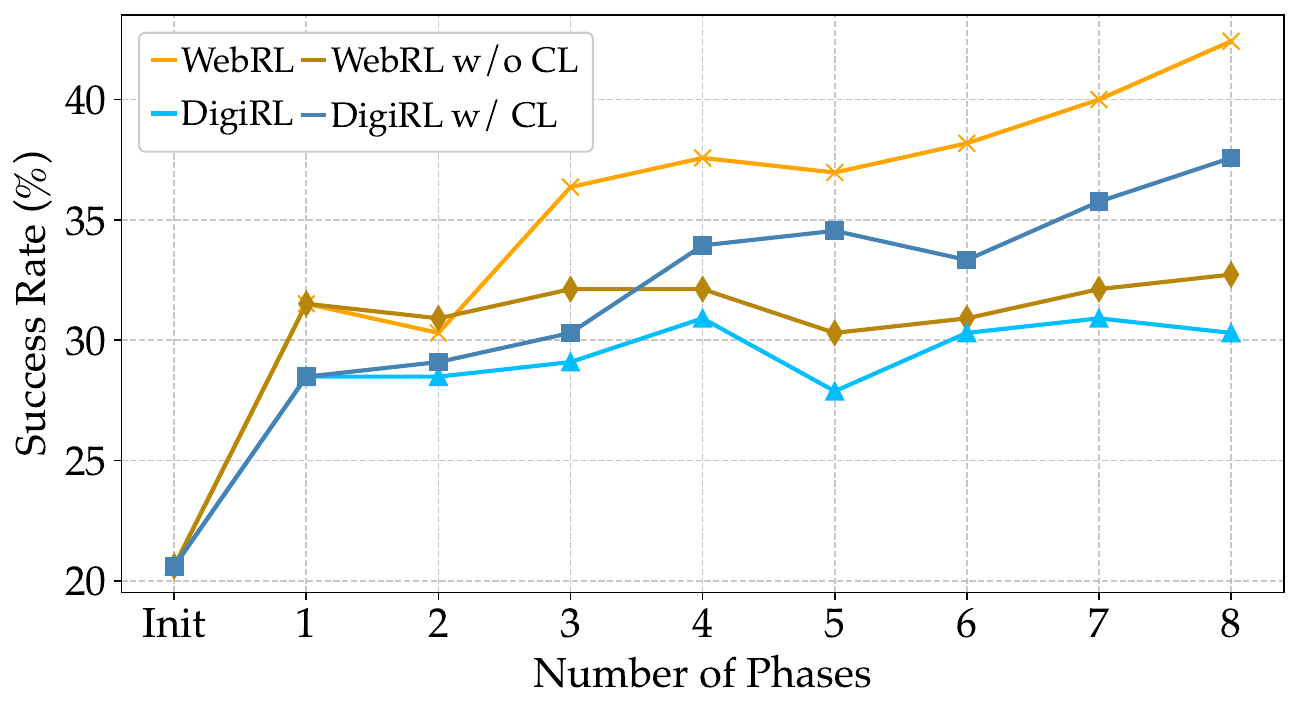}
    \vspace{-5mm}
    \caption{Comparison of \model and DigiRL with and without curriculum learning (CL).}
    \label{fig:ablation_w_digirl}
  \end{minipage}
\end{figure}

Figure~\ref{fig:error_bars} illustrates the error bars for \model across each phase. We conduct four sets of experiments, each utilizing a distinct random seed. The results from these four sets of experiments are presented in Figure~\ref{fig:error_bars}. Despite some fluctuations in accuracy, \model consistently demonstrates an overall upward trend. Furthermore, its accuracy consistently surpasses all baseline methods after completing the final phase of training.

Figure~\ref{fig:ablation_w_digirl} presents the success rate of \model and DigiRL, both with and without the application of our self-evolving curriculum learning. The results indicate that incorporating curriculum learning significantly enhances the performance of DigiRL. However, its performance still remains below that of \model. Conversely, when \model is implemented without curriculum learning, its performance experiences a notable decline, yet it still slightly surpasses that of DigiRL.

\begin{figure}[t]
  \centering
  \begin{minipage}[t]{0.48\textwidth}
    \centering
    \includegraphics[width=1.0\linewidth]{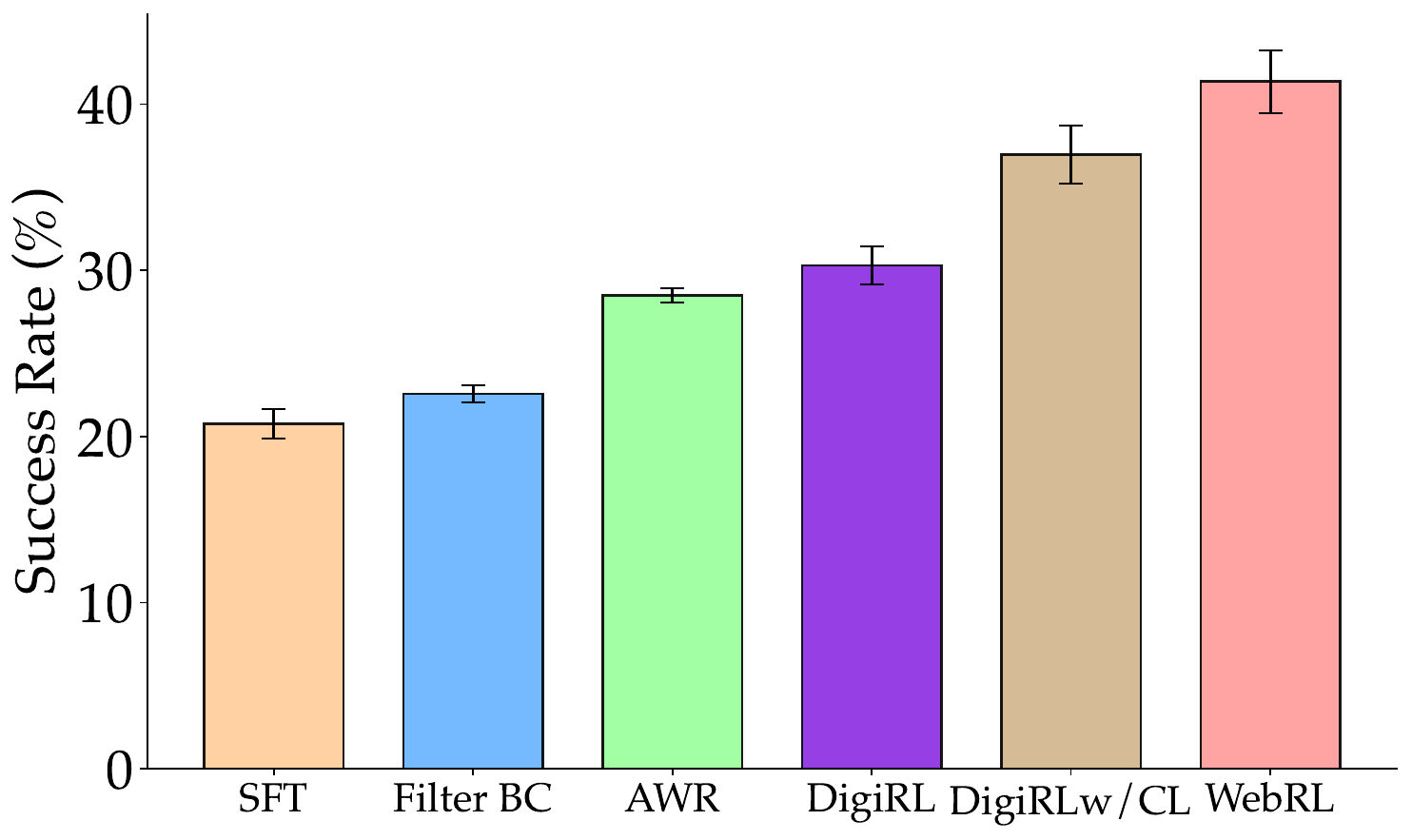}
    \vspace{-5mm}
    \caption{Error bars of baselines, DigiRL w/ CL, and \model.}
    \label{fig:error_bars_all}
  \end{minipage}
  \hfill
  \begin{minipage}[t]{0.50\textwidth}
    \centering
    \includegraphics[width=\linewidth]{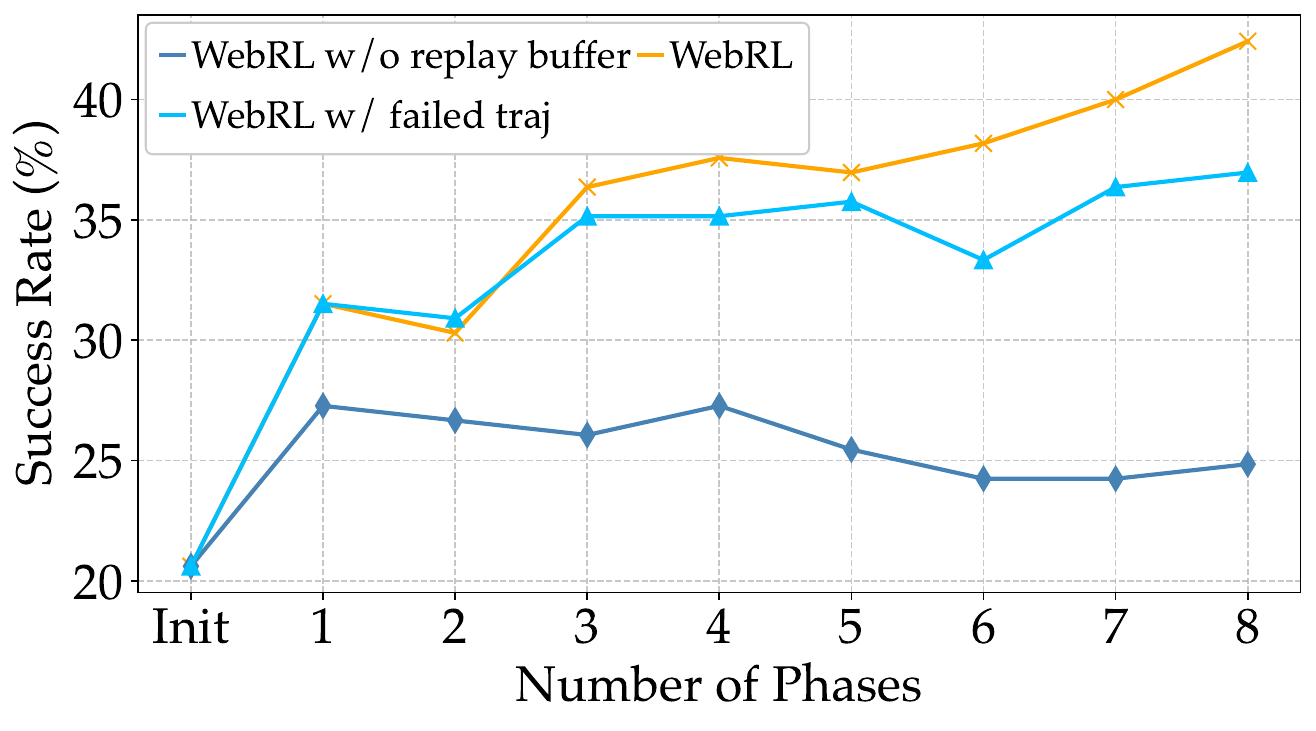}
    \vspace{-5mm}
    \caption{Comparison of \model and \model with failed trajectories in the replay buffer.}
    \label{fig:ablation_fail_traj}
  \end{minipage}
\end{figure}

Figure~\ref{fig:error_bars_all} illustrates the error bars for both the baselines, DigiRL w/ CL, and \model. Notably, the performance of \model and DigiRL with curriculum learning (CL) shows a higher standard deviation, which can be attributed to the inherent randomness in task generation. Despite this variability, \model consistently outperforms all baseline methods, demonstrating its effectiveness.

Figure~\ref{fig:ablation_fail_traj} compares two settings of the replay buffer: one containing only past successful trajectories and the other including all past trajectories, both successful and failed. The results show that including failed trajectories leads to slower and more volatile performance improvement. 
This issue arises because the advantage of an action in historical trajectories is estimated based on the trajectory's final reward. Including failed historical trajectories introduces a potential problem: even if certain actions in a failed trajectory are correct, they are likely to receive a negative advantage when the trajectory's final reward is zero. This reduces the likelihood of selecting these correct actions, thereby hindering effective policy improvement and leading to degraded performance.

\begin{wraptable}{r}{0.35\textwidth} 
    \centering
    \vspace{-2mm}
    \begin{threeparttable}
    \caption{Performance of different prompts for task generation.}
    \vspace{-2mm}
    \begin{tabularx}{0.35\textwidth}{>{\centering\arraybackslash}X*{2}{>{\centering\arraybackslash}X}}
        \toprule
        \text{Prompt 1} & \text{Prompt 2} & \text{Prompt 3}\\
        \midrule
        42.4 & 40.6 & 43.6 \\
        \bottomrule
    \end{tabularx}
    \label{tab:prompt}
    \end{threeparttable}
    \vspace{-3mm}
\end{wraptable}

We employ a variety of prompts to assess the stability of the curriculum learning strategy in \model. Specifically, we utilize three distinct prompts: Prompt 1 (in Figure~\ref{fig:prompt_for_task_generation}), Prompt 2 (in Figure~\ref{fig:prompt_for_task_generation2}), and Prompt 3 (in Figure~\ref{fig:prompt_for_task_generation3}). The training results for these prompts, measured after eight phases of training, are presented in Table~\ref{tab:prompt}. The results demonstrate that when using a variety of prompts to generate tasks, \model consistently delivers strong performance, highlighting the stability of our method.

%% file: section/appendix/prompts.tex
\section{Prompts Employed in WebRL}

\begin{figure}[ht]
    \centering
    \includegraphics[width=\linewidth]{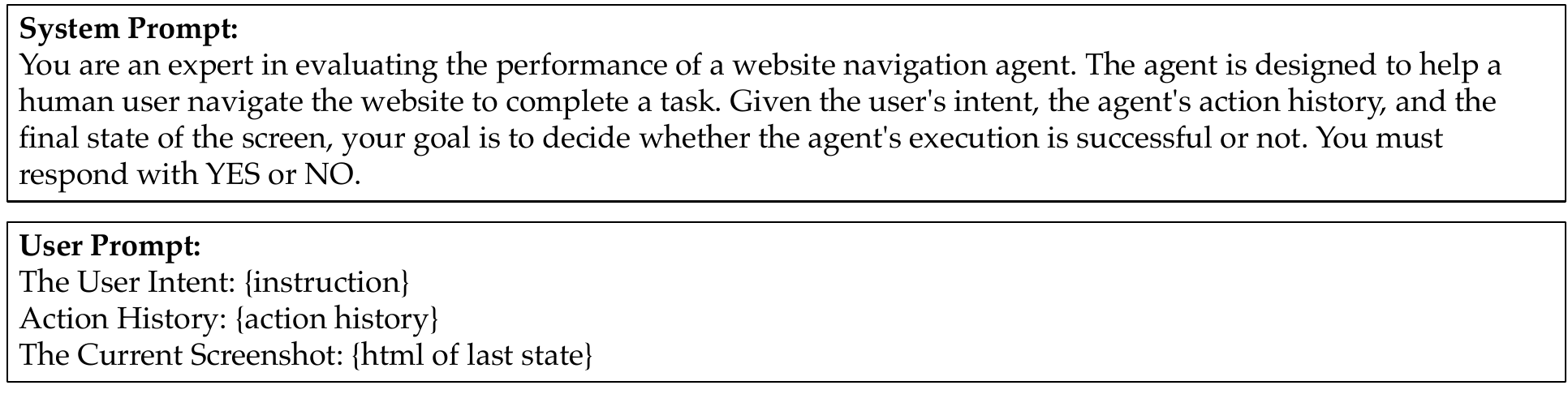}
    \caption{Prompts for \orm{} to assess the completion of Instructions.}
    \label{fig:prompt_for_reward_model}
\end{figure}

The prompt used for \orm{} is shown in Figure~\ref{fig:prompt_for_reward_model}. We require the model to output ``YES'' or ``NO'' to determine whether a certain trajectory successfully completes the corresponding task. Considering the limited context size, we only input the HTML content of the last state.

\label{sec:prompts}

\begin{figure}[htbp]
    \centering
    \includegraphics[width=\linewidth]{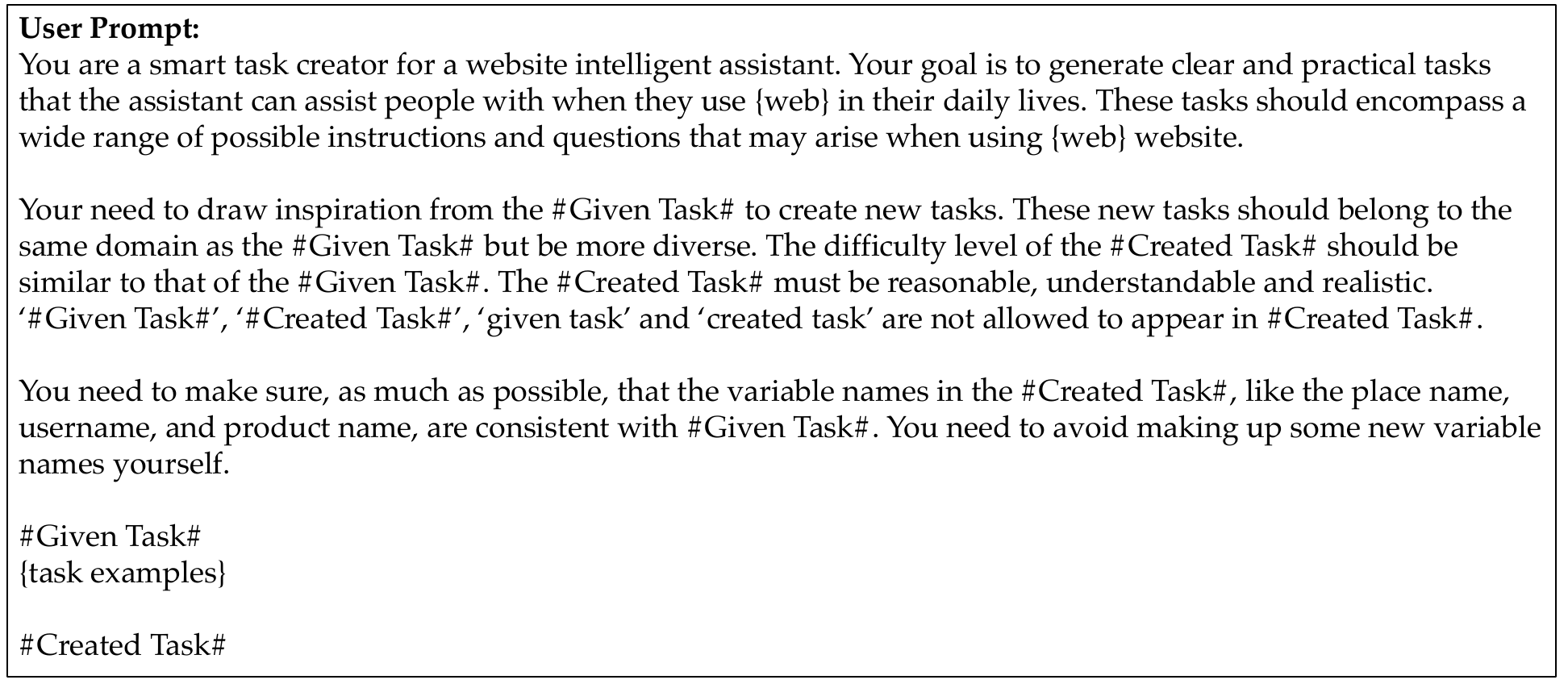}
    \caption{Prompts for instruction generation.}
    \label{fig:prompt_for_task_generation}
\end{figure}

\begin{figure}[htbp]
    \centering
    \includegraphics[width=\linewidth]{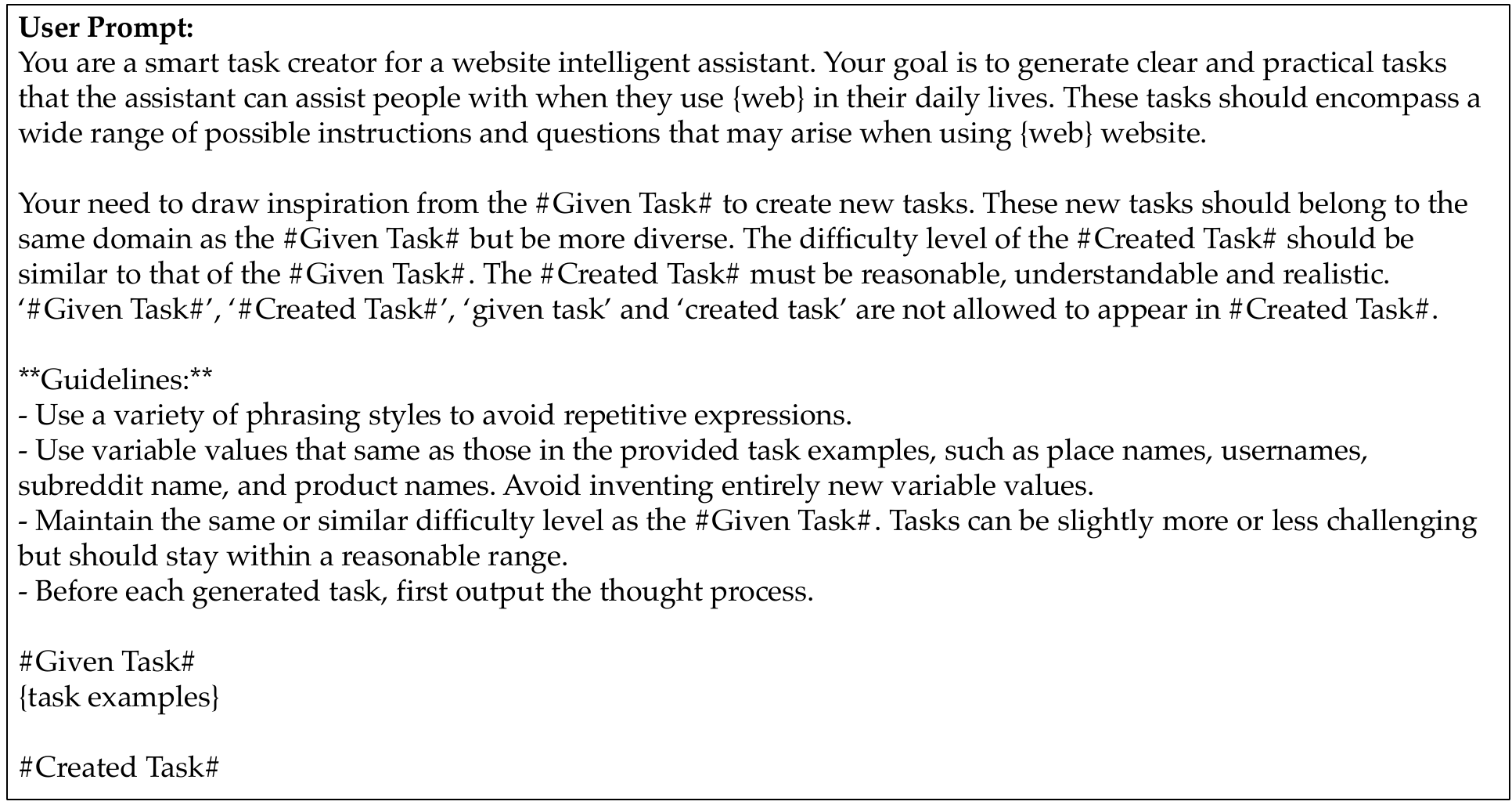}
    \caption{Prompts for instruction generation (version 2).}
    \label{fig:prompt_for_task_generation2}
\end{figure}

\begin{figure}[htbp]
    \centering
    \includegraphics[width=\linewidth]{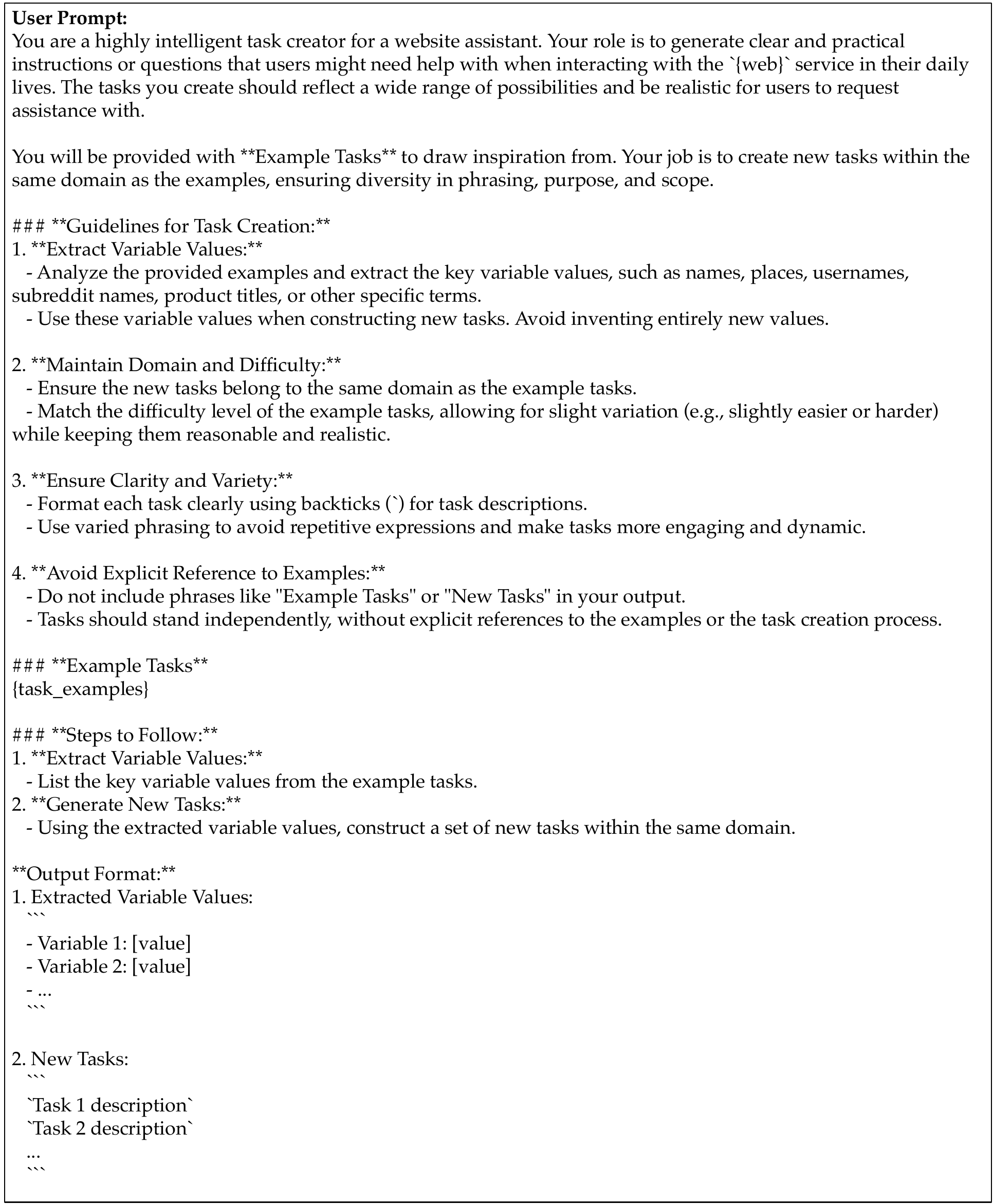}
    \caption{Prompts for instruction generation (version 3).}
    \label{fig:prompt_for_task_generation3}
\end{figure}

The prompt for generating new instructions is presented in Figure~\ref{fig:prompt_for_task_generation}. We use tasks that the model fails to complete successfully in previous phases as seeds for generating new tasks of similar difficulty but with greater variety. The prompts shown in Figure~\ref{fig:prompt_for_task_generation2} and Figure~\ref{fig:prompt_for_task_generation3} represent alternative versions of the prompts used for task generation.

\begin{figure}[ht]
    \centering
    \includegraphics[width=\linewidth]{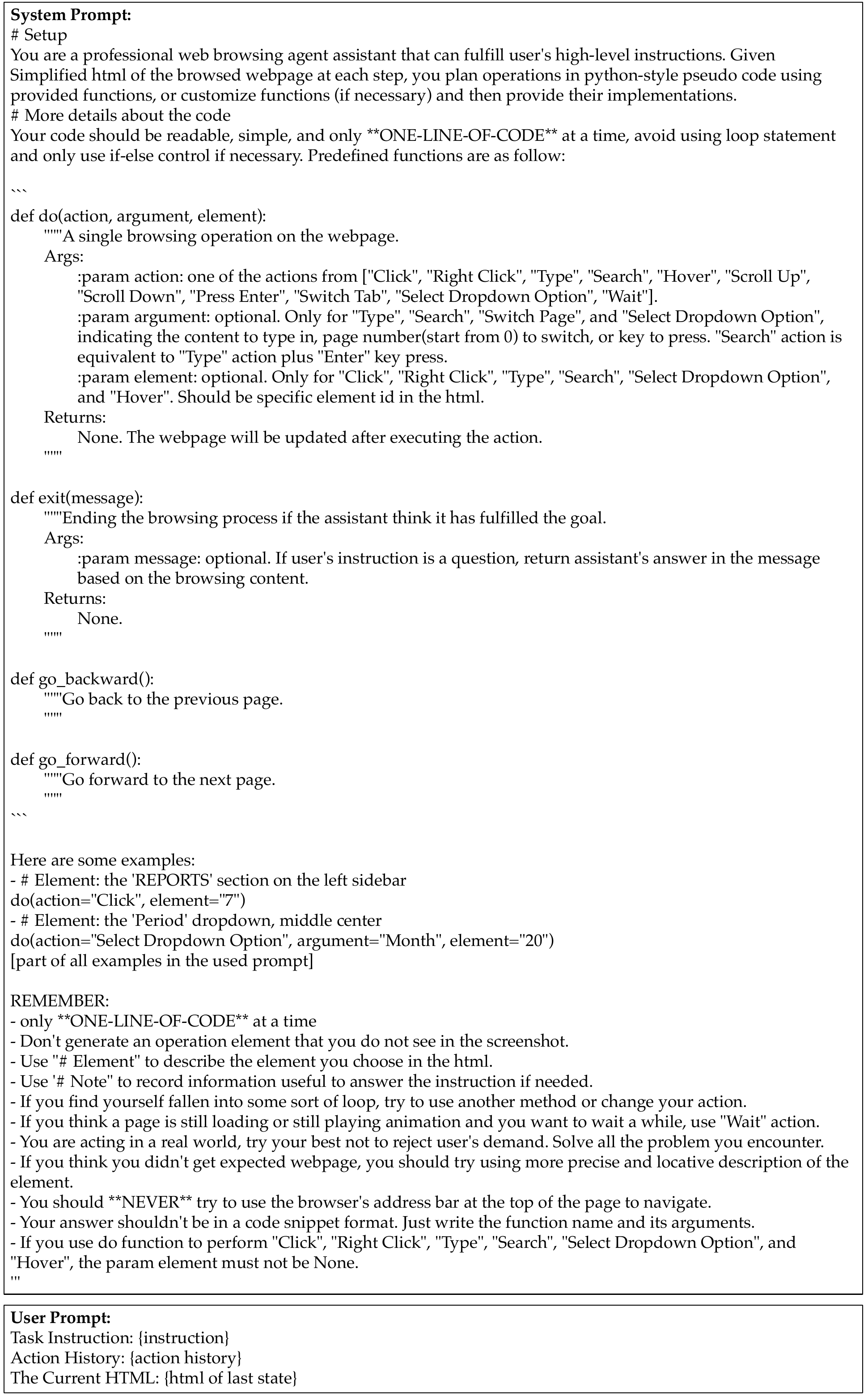}
    \caption{The simple prompt employed in baselines.}
    \label{fig:prompt_for_baseline}
\end{figure}

The simple prompt we use to test models including GPT-4-Turbo, GPT-4o, Llama3.1-8B-Instruct, and Llama3.1-70B-Instruct is shown in Figure~\ref{fig:prompt_for_baseline}. In this prompt, we define the feasible actions and provide illustrative examples. Additionally, we outline a set of requirements in the ``REMEMBER'' part to guide the model's behavior.

\begin{figure}[ht]
    \centering
    \includegraphics[width=\linewidth]{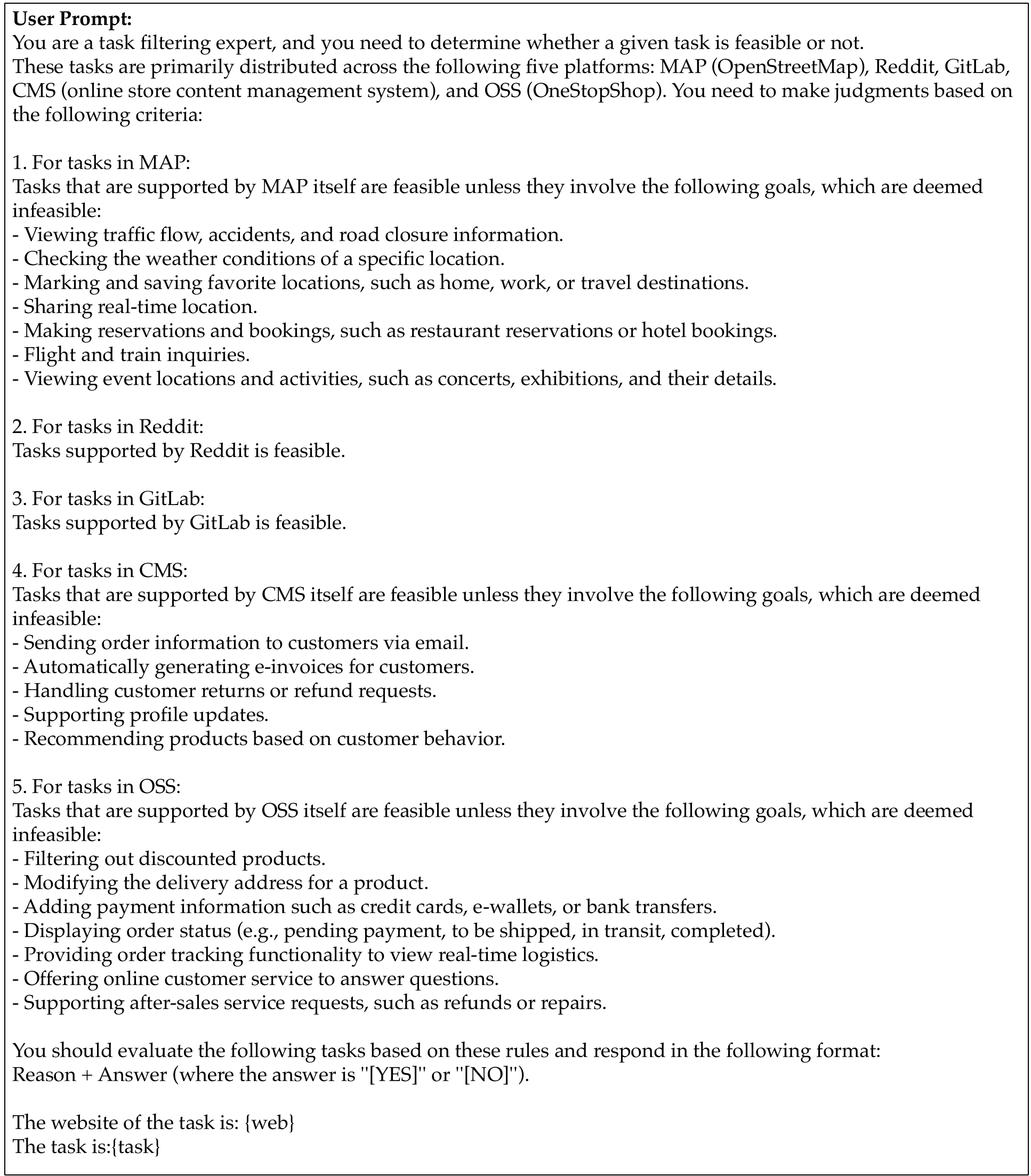}
    \caption{The prompt used to filter instructions.}
    \label{fig:prompt_for_filter}
\end{figure}

The prompt we use to filter the instructions generated by GPT-4o is presented in Figure~\ref{fig:prompt_for_filter}. In this prompt, we provide a rule defining the feasibility of instructions across the five websites in WebArena.

%% file: section/appendix/qualitive.tex
\clearpage
\section{Qualitative Examples}
\label{sec:examples}

We list one example of \model on each of the five sites in WebArena.

\begin{figure}[ht]
    \centering
    \includegraphics[width=\linewidth]{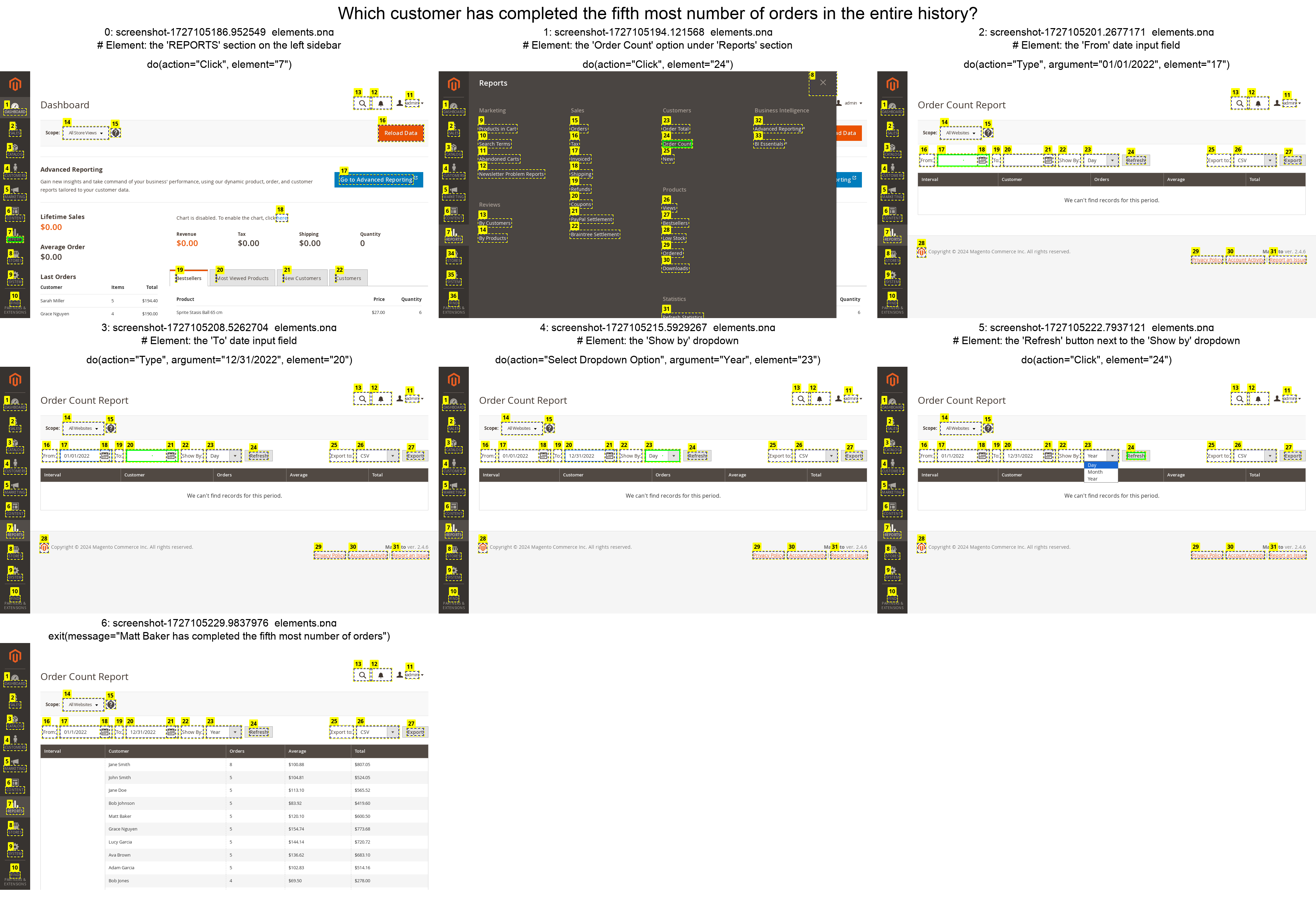}
    \caption{CMS Example.}
\end{figure}

\begin{figure}[ht]
    \centering
    \includegraphics[width=\linewidth]{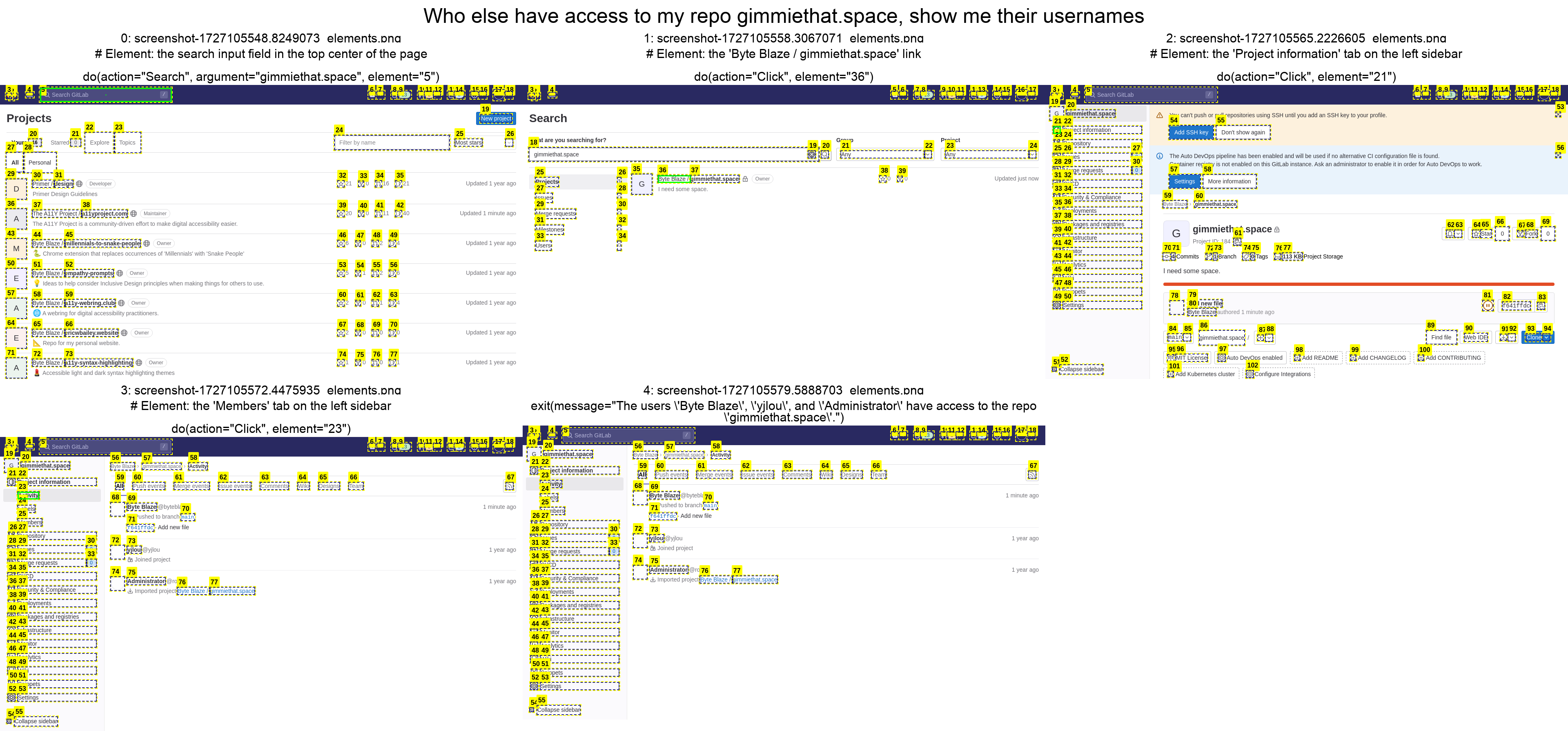}
    \caption{Gitlab Example.}
\end{figure}

\begin{figure}[ht]
    \centering
    \includegraphics[width=\linewidth]{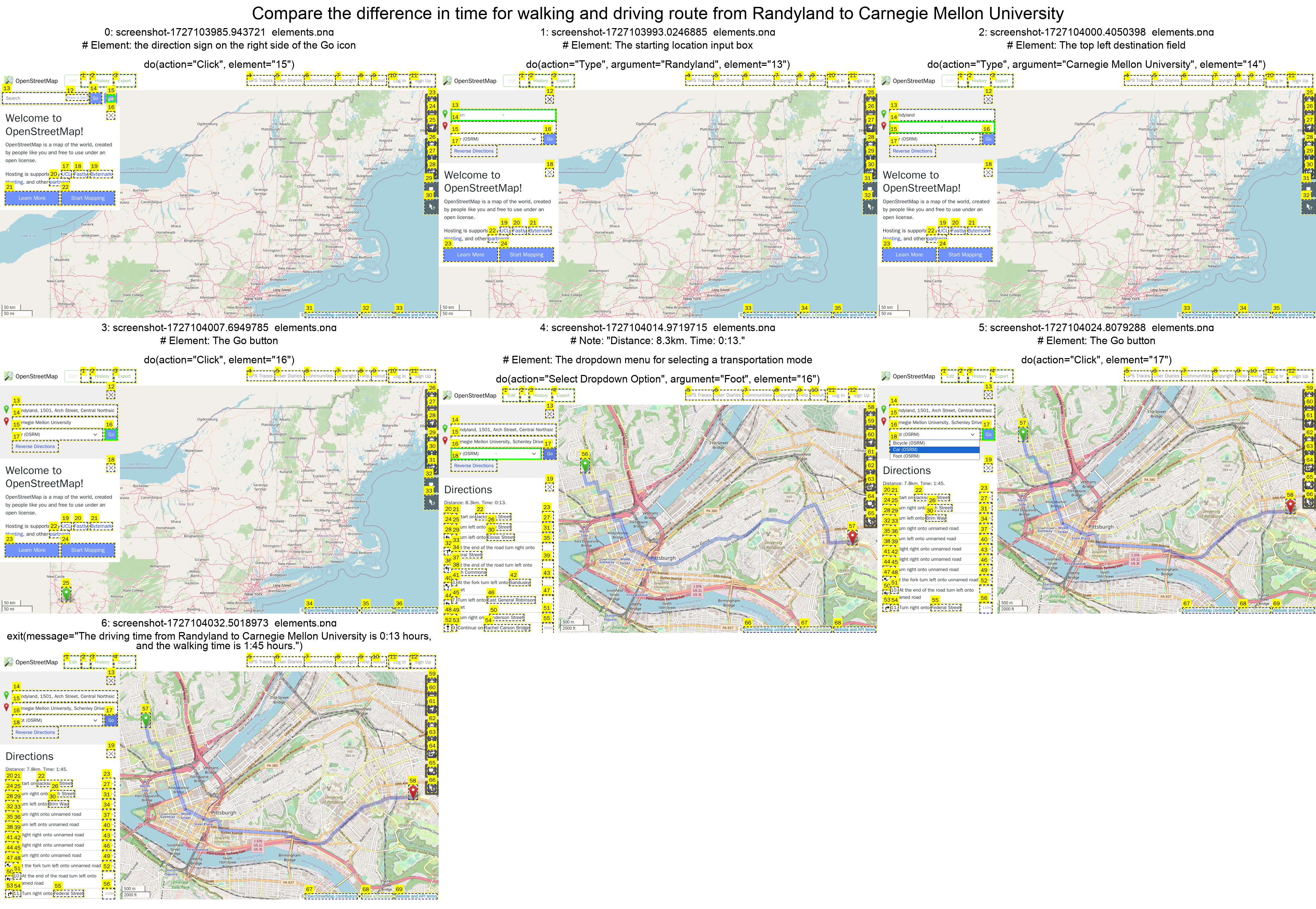}
    \caption{MAP Example.}
\end{figure}

\begin{figure}[ht]
    \centering
    \includegraphics[width=\linewidth]{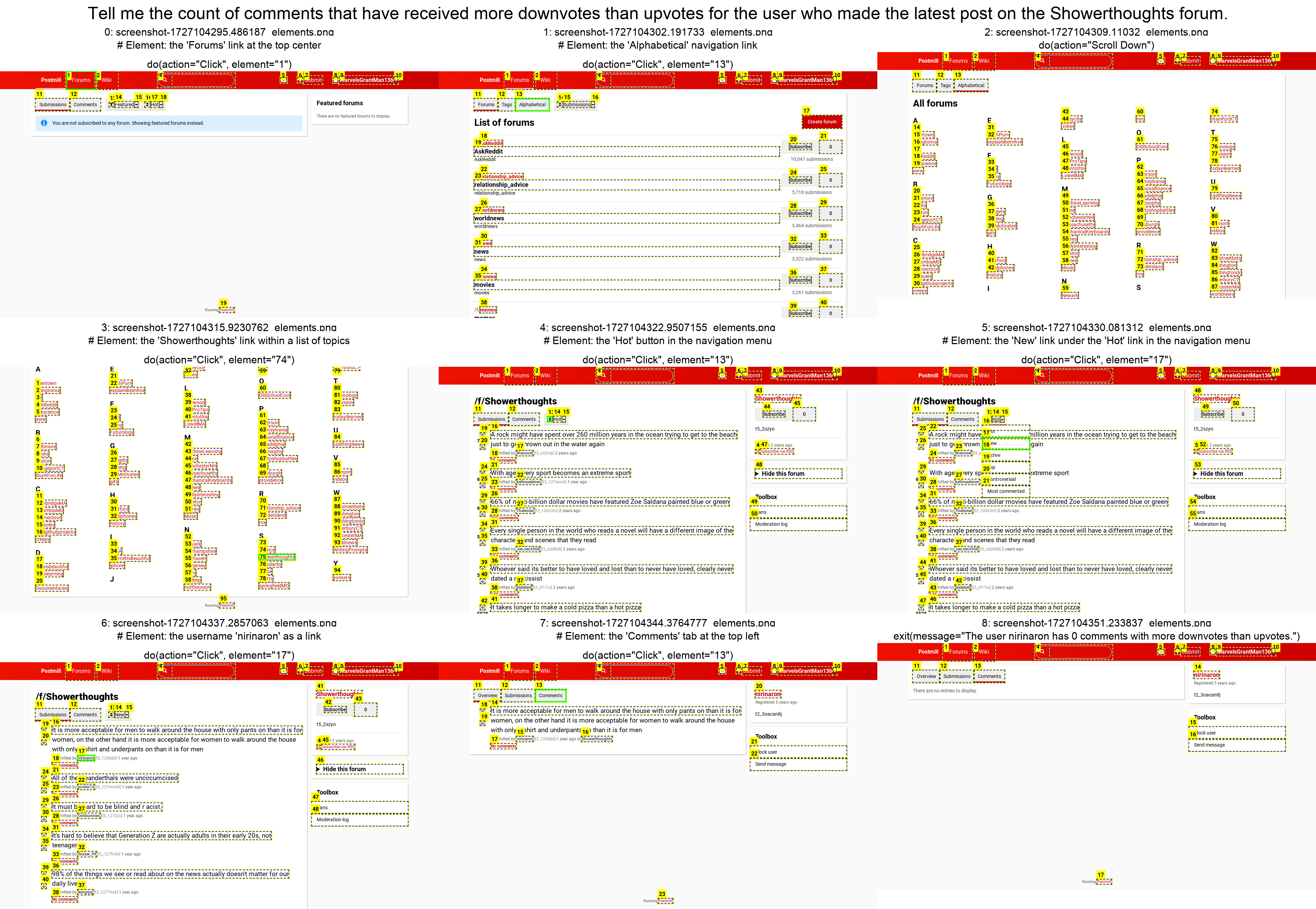}
    \caption{Reddit Example.}
\end{figure}

\begin{figure}[ht]
    \centering
    \includegraphics[width=\linewidth]{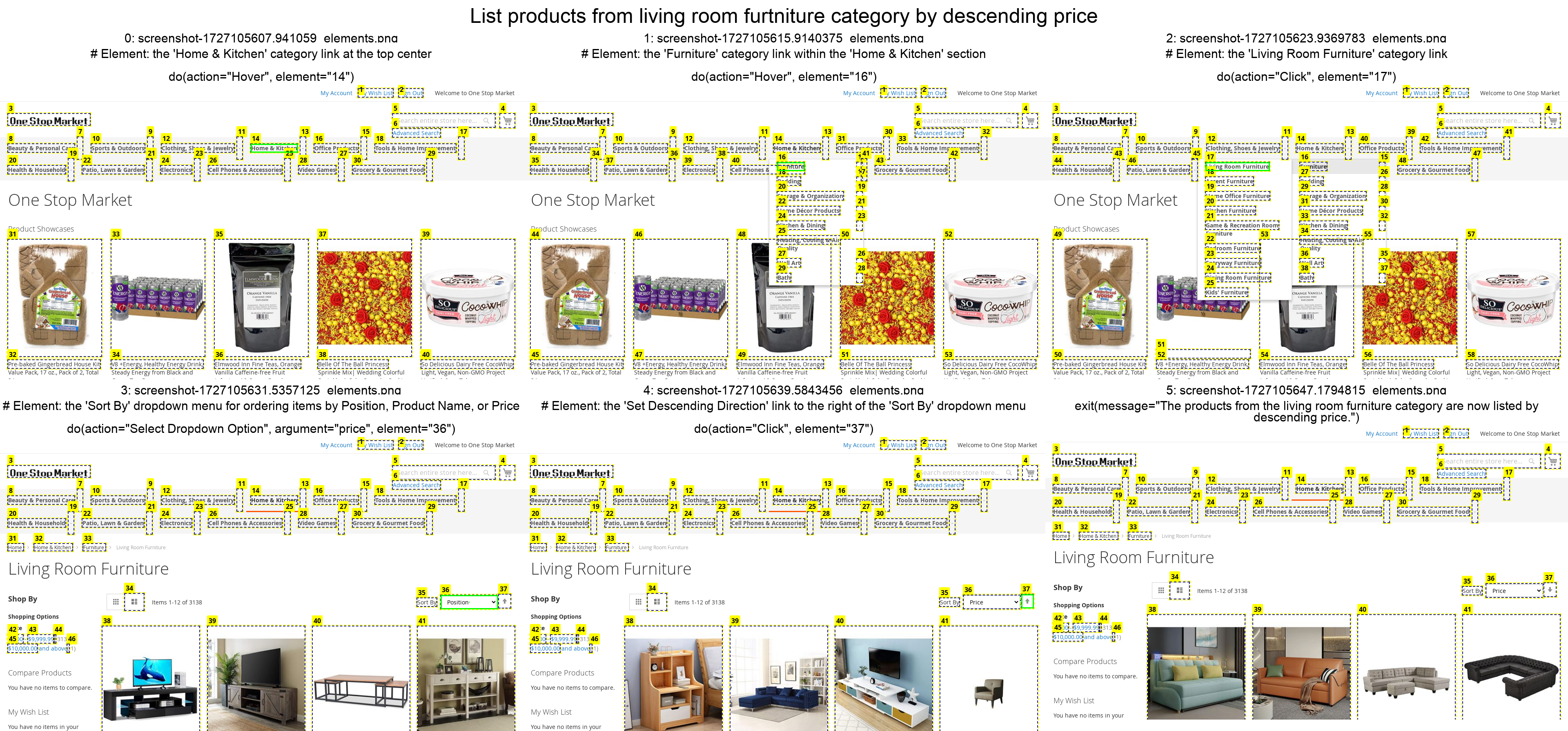}
    \caption{OSS Example.}
\end{figure}